%% file: main.tex
\RequirePackage[svgnames]{xcolor}

\documentclass[11pt,letterpaper]{mystyle}

\usepackage[all]{hypcap}
\usepackage[svgnames]{xcolor}
\usepackage[numbers,sort&compress]{natbib}
\usepackage{hyperref}[citecolor=lightblue]

\hypersetup{
    colorlinks = true,
    citecolor = {YaleBlue},
}

\usepackage{algorithm}
\usepackage{algorithmicx}
\usepackage{algpseudocode}
\usepackage{microtype}
\usepackage{graphicx}
\expandafter\def\csname ver@subfig.sty\endcsname{}
\usepackage{booktabs} %
\usepackage{float}
\usepackage{bigstrut}

\usepackage{amsmath}
\usepackage{amssymb}
\usepackage{mathtools}
\usepackage{amsthm}
\usepackage{mathrsfs}
\usepackage{nicefrac}
\usepackage{dsfont}
\usepackage{enumitem}
\usepackage{subcaption}
\usepackage{graphicx,subfig}
\usepackage{cleveref}
\usepackage{bxcoloremoji}

\usepackage{float}

\usepackage[utf8]{inputenc} %
\usepackage[T1]{fontenc}    %
\usepackage{hyperref}       %
\usepackage{url}            %
\usepackage{booktabs}       %
\usepackage{amsfonts}       %
\usepackage{nicefrac}       %
\usepackage{microtype}      %
\usepackage{graphicx}
\usepackage{subcaption} 
\usepackage{wrapfig}
\usepackage{lipsum}
\usepackage{enumitem}
\usepackage{stackengine}
\usepackage[font=small,labelfont=bf]{caption}
\usepackage{color}
\usepackage{adjustbox}

\usepackage{rotating}
\usepackage{makecell}

\usepackage{orcidlink}
\usepackage{multirow} 
\usepackage{bm} 
\usepackage{xcolor} 
\usepackage{colortbl} 
\usepackage{caption} 
\usepackage{pifont} 
\usepackage{algorithm}
\usepackage{algpseudocode}
\usepackage{amsfonts}
\usepackage{soul} 
\usepackage{enumitem}
\usepackage{listings}

\tcbuselibrary{skins, breakable}

\definecolor{headerGray}{RGB}{85, 85, 85}
\definecolor{varBlue}{RGB}{0, 0, 200}
\definecolor{codeBg}{RGB}{245, 245, 245}

\newcommand{\var}[1]{\textcolor{varBlue}{\texttt{\{#1\}}}}

\input{macro}

\newtcolorbox{AIbox}[2][]{aibox,title=#2,#1}
\definecolor{lightblue}{rgb}{0.22,0.45,0.70}%
\definecolor{Gray}{gray}{0.95}
\definecolor{Cornsilk}{rgb}{1.0, 0.97, 0.86}

\newcommand{\mypara}[1]{\smallskip\noindent\textbf{#1}}

\crefname{section}{Sec.}{Secs.}
\Crefname{section}{Sec.}{Secs.}
\crefname{subsection}{Sec.}{Secs.}
\Crefname{subsection}{Sec.}{Secs.}

\newcommand{\cmark}{\ding{51}} 
\newcommand{\xmark}{\ding{55}} 

\newcommand{\apptocA}[2]{%
  \noindent\hyperref[#1]{\textbf{#2}}%
  \leaders\hbox{\kern.35em.\kern.35em}\hfill%
  \pageref{#1}\par
}
\newcommand{\apptocB}[2]{%
  \noindent\hspace*{2.0em}\hyperref[#1]{#2}%
  \leaders\hbox{\kern.35em.\kern.35em}\hfill%
  \pageref{#1}\par
}

\usepackage{amsmath}

\usepackage[all]{hypcap}

\title{Scaling Multi-Reference Image Generation with Dynamic Reward Optimization}

\runningtitle{Scaling Multi-Reference Image Generation with Dynamic Reward Optimization (ECCV2026)}

\author{
  Wenwang Huang$^{1,*}$,
  Yusen Fu$^{1,*}$,
  Junjie Wang$^{1}$,
  Mengfei Huang$^{1}$,
  Yulin Li$^{1}$,
  Gan Liu$^{2}$,
  Jing Cai$^{2}$,
  Yancheng He$^{2}$, and
  Zhuotao Tian$^{1,3,\dagger}$
}

\affil[1]{Harbin Institute of Technology, Shenzhen}
\affil[2]{Independent Researcher}
\affil[3]{Shenzhen Loop Area Institute}

\correspondingauthor{ Zhuotao Tian$\dagger$, \href{mailto:tianzhuotao@hit.edu.cn}{tianzhuotao@hit.edu.cn};
\quad
$*$ Equal contribution}

\begin{document}

\input{sections/abstract}

\maketitle

\vspace{3mm}
\input{sections/introduction}
\input{sections/dataset}
\input{sections/method}

\input{sections/experiments}
\input{sections/conclusion}
\clearpage
\bibliography{main}

\appendix
\input{sections/appendix}
\appendix
\end{document}

%% file: macro.tex
\newcommand{\x}{\mathbf{x}}

\definecolor{blanchedalmond}{rgb}{1.0, 0.92, 0.8}
\definecolor{carmine}{rgb}{0.59, 0.0, 0.09}
\definecolor{lightblue}{rgb}{0.22,0.45,0.70}%

\renewcommand{\mathbf}{\boldsymbol}

\makeatletter
\def\Ddots{\mathinner{\mkern1mu\raise\p@
\vbox{\kern7\p@\hbox{.}}\mkern2mu
\raise4\p@\hbox{.}\mkern2mu\raise7\p@\hbox{.}\mkern1mu}}
\makeatother

\definecolor{amaranth}{rgb}{0.9, 0.17, 0.31}
\definecolor{antiquebrass}{rgb}{0.8, 0.58, 0.46}
\definecolor{antiquefuchsia}{rgb}{0.57, 0.36, 0.51}
\definecolor{chromeyellow}{rgb}{0.31, 0.47, 0.26}

\newcommand{\github}{\raisebox{-1.5pt}{\includegraphics[height=1.05em]{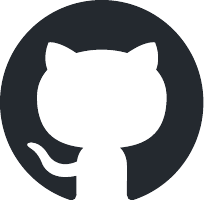}}}

%% file: sections/abstract.tex
\begin{abstract}
While personalized image generation has achieved remarkable progress, multi-reference image generation (MRIG) remains a challenging task. Most existing benchmarks fail to adequately evaluate complex MRIG scenarios, hindering further progress in this area. To better assess model performance on complex MRIG tasks, we introduce OmniRef-Bench, a benchmark that covers complex combinations of reference image types and a large number of reference images. Evaluations on OmniRef-Bench show that mainstream open-source models struggle in complex MRIG scenarios, and their performance deteriorates significantly as the number of mixed-type reference images increases.
To address this issue, we propose \textbf{DyRef}, a two-stage training framework. In the first stage, supervised fine-tuning equips the model with the basic capability to handle complex MRIG tasks. In the second stage, we introduce Difficulty-aware Advantage Reweighting (DAR) and Discriminative Reward Scaling (DRS). DAR dynamically adjusts the optimization objective to improve performance when handling a large number of mixed-type reference images. DRS enlarges intra-group reward differences for more effective policy optimization. Experiments demonstrate that DyRef significantly improves the performance of open-source models on OmniRef-Bench and single-image editing benchmarks, demonstrating the effectiveness and generalization capability of our approach.

\vspace{2mm}


\vspace{5mm}



\github{} \textbf{Code Repository}: \href{https://github.com/Weistrass/DyRef}{https://github.com/Weistrass/DyRef}



\coloremojicode{1F4DA} \textbf{Datasets}: \href{https://huggingface.co/datasets/Eason0438/OmniRef-training}{https://huggingface.co/datasets/Eason0438/OmniRef-training}

\coloremojicode{1F3C6} \textbf{Benchmark}: \href{https://huggingface.co/datasets/Eason0438/OmniRef-Bench}{https://huggingface.co/datasets/Eason0438/OmniRef-Bench}

\coloremojicode{1F4E7} \textbf{Contact}: \href{mailto:liamwwhuang@163.com}{liamwwhuang@163.com}

\end{abstract}

%% file: sections/introduction.tex
\vspace{-4mm}
\section{Introduction}
\begin{figure*}[t]
\centering
\includegraphics[width=0.9\textwidth]{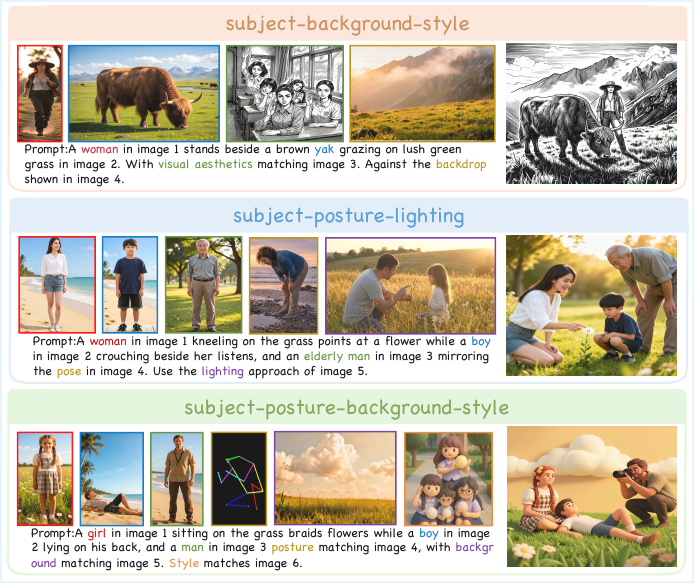}
\caption{\textbf{Versatile samples of the proposed DyRef}. Given complex samples that involve mixed reference image types and varying numbers of reference images, DyRef consistently generates high-quality results in accordance with user instructions. }
\label{fig:figure1_show}
\end{figure*}
The emergence of diffusion models~\cite{peebles2023scalable,rombach2022high} has substantially advanced personalized image generation~\cite{wu2025qwenimagetechnicalreport,liu2025step1x,flux-2-2025,mou2025dreamo,tan2025ominicontrol}, enabling high-fidelity synthesis closely aligned with user instructions. However, 
among various personalization paradigms,
Multi-Reference Image Generation (MRIG)~\cite{xia2025dreamomni3,wei2026uniref,xia2025dreamomni2,song20253sgen,oshima2025multibanana} remains challenging due to the difficulty of effectively integrating multiple reference images of diverse types (e.g., subject, pose, and style). Addressing these challenges carries significant implications for fields like professional visual design~\cite{wang2025generative} and advertising applications~\cite{chen2025ctr}, where coherent integration of varied visual elements is crucial.

\mypara{Insufficient Evaluation for Complex MRIG.}
Despite its potential, MRIG research remains constrained by insufficient benchmarks. Existing benchmarks~\cite{chen2025xverse,chen2025multiref,oshima2025multibanana,xu2025contextgen} typically involve simple tasks with limited numbers and types of reference images, neglecting complex scenarios that require integrating numerous and diverse references. Additionally, these benchmarks lack fine-grained and multi-dimensional evaluation metrics to accurately assess model performance across various reference types. These limitations hinder meaningful comparisons between methods and impede progress towards practical MRIG applications.

\begin{figure*}[t]
\centering
\includegraphics[width=0.9\linewidth]{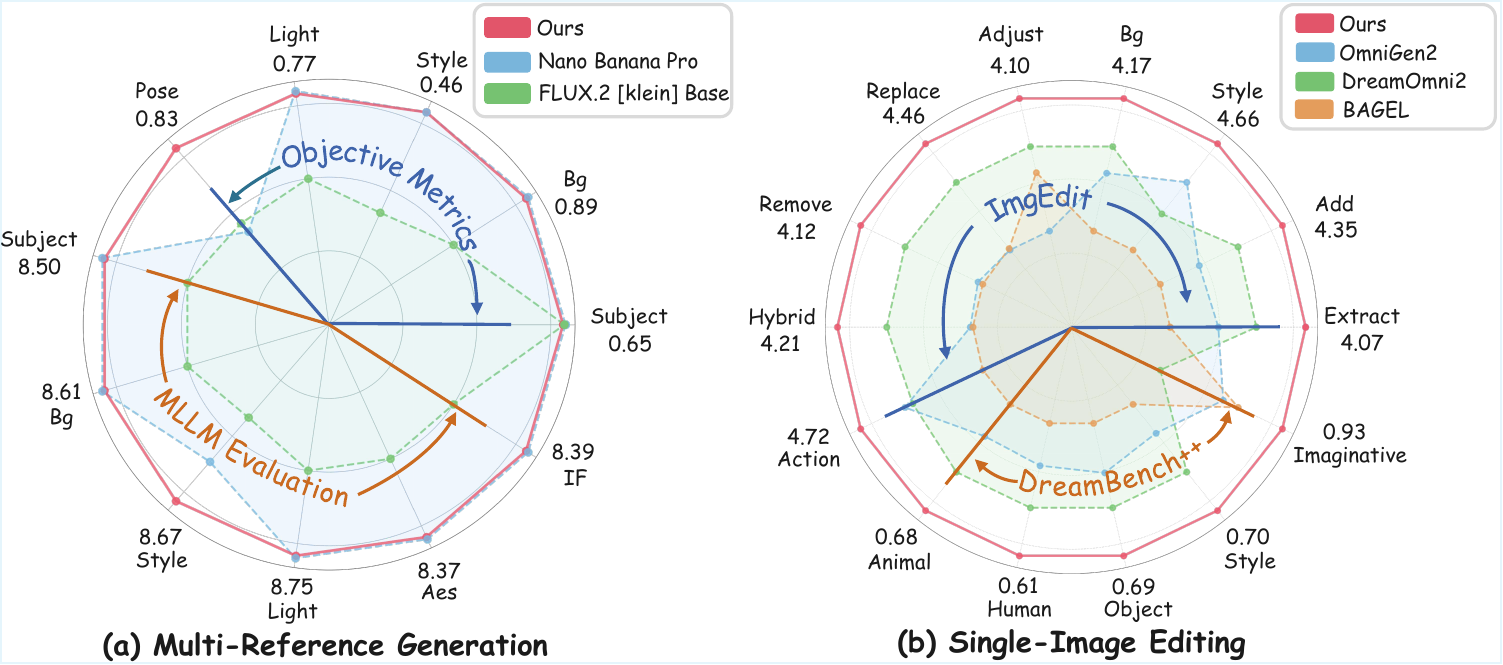}
\caption{\textbf{Performance of DyRef}. On both MRIG and single-image editing tasks, DyRef (Ours) consistently outperforms open-source state-of-the-art methods and achieves performance comparable to leading closed-source model Nano Banana Pro.}
\label{fig:radar plot}
\end{figure*}

\mypara{The Proposed OmniRef-Bench.}
To better assess MRIG performance, we introduce OmniRef-Bench, a personalized image generation benchmark comprising intricate combinations of diverse reference types as well as a large number of reference images across synthetic and real-world scenarios. Furthermore, we propose a hybrid evaluation strategy combining objective quantitative metrics with Multimodal Large Language Models (MLLMs)-based assessments, and design fine-grained evaluation criteria tailored to each reference type, facilitating a robust and multidimensional evaluation of complex MRIG tasks.

\begin{figure*}[t]
\centering
\includegraphics[width=\linewidth]{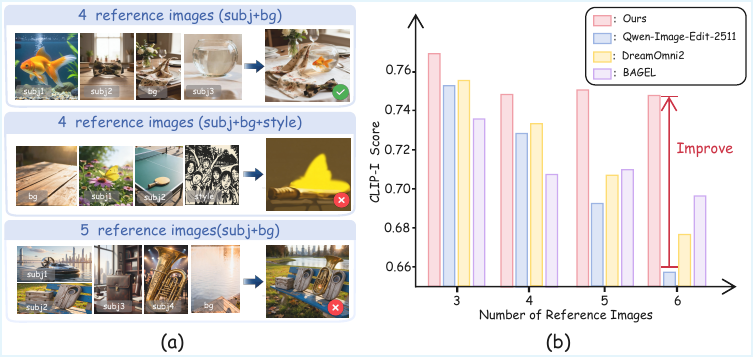}
\caption{\textbf{Motivation of our DyRef}. \textbf{(a)} Qualitative results: open source models yield high-quality results with a limited set of references, but suffer from significant artifacts and semantic loss as the number and complexity of reference images increase. \textbf{(b)} Quantitative results: on the OmniRef-Bench, CLIP-I scores between generated images and target images confirm that the performance of mainstream open-source models significantly drops as the number of mixed-type reference images increases.}
\label{fig:method_motivation}
\end{figure*}

\mypara{Key Observations.}
Using OmniRef-Bench (Tab.~\ref{tab:results_Omni-Ref}), we find that mainstream open-source MRIG models~\cite{wu2025qwenimagetechnicalreport,xia2025dreamomni2,deng2025bagel,wu2025omnigen2} struggle in complex MRIG scenarios, exhibiting a huge gap compared with state-of-the-art closed-source models~\cite{Nanobanana,seedream2025seedream}. To bridge this disparity, we conduct further experiments and observe that while open-source models generate high-quality images with a few references, their performance deteriorates rapidly as the number of reference images involving mixed reference types increases, as illustrated in Fig.~\ref{fig:method_motivation}. This observation raises a critical research question: \textit{Can mitigating this degradation effectively narrow the gap between open-source and closed-source models?}

\mypara{Our Solution.}
To address this issue, we propose \textbf{DyRef}, a two-stage training framework. In the first stage, we use Supervised Fine-Tuning (SFT) to equip the model with preliminary capabilities to handle complex MRIG tasks. In the second stage, we introduce Difficulty-aware Advantage Reweighting (DAR) and Discriminative Reward Scaling (DRS).
Specifically, DAR dynamically adjusts the optimization objective to encourage the model to focus more on underperforming samples with a larger number of mixed-type reference images, while preventing overfitting to well-performing samples with fewer references. DRS enlarges reward differences within each group for more effective policy optimization.
The combination of DAR and DRS enhances the model’s ability to learn from samples with a large number of diverse reference images, thereby improving performance on complex MRIG tasks.

Experimental results (Fig.~\ref{fig:figure1_show}, Fig.~\ref{fig:radar plot} and Fig.~\ref{fig:method_motivation}(b)) demonstrate that DyRef substantially improves the performance of open-source models on OmniRef-Bench, achieving results comparable to state-of-the-art closed-source models. Moreover, DyRef further enhances the model performance on single-image editing benchmarks, highlighting its effectiveness and generalization ability.

In summary, the contributions of this work are as follows:
\begin{itemize}[label=$\bullet$]

    \item To better assess model performance on complex MRIG tasks, we introduce OmniRef-Bench. Using this benchmark, we observe that open-source models struggle in complex multi-reference settings, and their performance declines rapidly as the number of mixed-type reference images increases.

    \item To address the issue, we propose DyRef. Building upon SFT, we employ DAR to encourage the model to focus more on underperforming samples with a larger number of mixed-type reference images, and utilize DRS to enhance reward differences across samples for better policy optimization.

    \item Experimental results show that DyRef not only substantially boosts performance on OmniRef-Bench, but also improves results on single-image editing benchmarks, demonstrating both its effectiveness and generalization.
\end{itemize}

\label{sec:intro}
\vspace{-1mm}

%% file: sections/dataset.tex
\section{Dataset and Benchmark}
\label{sec:datasets}
In this section, we first introduce the motivation of OmniRef-Bench in~\Cref{subsec:bench_motivation}. Next,~\Cref{subsec:benchmark} describes the composition and construction pipeline of the benchmark. Finally,~\Cref{subsec:bench_eval} outlines the corresponding evaluation strategy.
\begin{table*}[t]
\centering
\setlength{\tabcolsep}{8pt}  
\renewcommand{\arraystretch}{1.1}
\caption{Comparison of OmniRef-Bench with existing benchmarks. Eval Granularity indicates whether a benchmark conducts tailored assessment for every reference type (fine-grained), or simply conducts a general evaluation (coarse-grained). Max Combs denotes the maximum number of reference image type combinations. Obj refers to objective quantitative metrics. MLLM indicates evaluation conducted using the MLLM.}
\begin{tabular}{l|cccl}
\toprule
Benchmarks    & Max Combs. & Expert-Annotated & Eval Granularity & Eval Perspective             \\ \midrule
XVerse~\cite{chen2025xverse}        & 1                                     & \xmark             & fine-grained          & Obj      \\
PSRBench~\cite{wang2025psr}           &1                              & \xmark                  & fine-grained     & Obj + MLLM                    \\
MultiBanana~\cite{oshima2025multibanana}         & 3                                   & \xmark   & coarse-grained            & MLLM                     \\
DreamOmni2~\cite{xia2025dreamomni2}           & 2                                   & \cmark        & coarse-grained                 & MLLM            \\
OmniRef-Bench (Ours)       & 4                            & \cmark   & fine-grained              & Obj + MLLM \\ \bottomrule
\end{tabular}
\label{tab:bench_comparision}
\end{table*}
\subsection{Motivation}
\label{subsec:bench_motivation}
As shown in Tab.~\ref{tab:bench_comparision}, existing benchmarks exhibit limitations in evaluating complex MRIG tasks. On the one hand, most existing benchmarks~\cite{chen2025xverse,peng2024dreambench,wang2025psr} primarily focus on single-subject or multi-subject generation. Although some consider a large number of reference images, they lack combinations of multiple reference types, such as style, lighting, and pose. On the other hand, recent studies~\cite{xu2025contextgen,chen2025multiref,xia2025dreamomni2} have attempted to incorporate reference types beyond subject identity, but they typically only consider simple pairwise combinations with a small number of reference images, which fail to satisfy the complex semantic compositions required for artistic creation and advanced content generation.

Moreover, existing benchmarks~\cite{song20253sgen,ye2025imgedit} typically rely exclusively on either objective quantitative metrics or MLLM-based scoring mechanisms. However, relying on only one of these approaches introduces inherent limitations. Detailed limitations are provided in Appendix~\ref{subsec:comparison of obj and mllm}. Furthermore, some benchmarks~\cite{oshima2025multibanana,xia2025dreamomni3} do not provide fine-grained evaluation criteria for each reference type, making it difficult to accurately measure model performance under combinations of diverse reference types. To address these limitations, we introduce OmniRef-Bench.

\subsection{OmniRef-Bench}
\label{subsec:benchmark}

\mypara{Benchmark Overview.}
OmniRef-Bench comprises a total of 395 expert-annotated benchmark instances. These instances cover five major reference types: subject, background, style, lighting, and pose. Each instance contains a target editing task together with multiple reference images, with the number of reference images ranging from 2 to 7. This design enables comprehensive evaluation from simple two-reference editing to complex MRIG. In addition, OmniRef-Bench is curated to include flexible and complex combinations of reference types, totaling 10 distinct combinations, with up to 4 reference types present simultaneously. More statistics about OmniRef-Bench can be found in Appendix~\ref{subsec:statistic_bench}.

\begin{figure*}[tbp]
\centering
\includegraphics[width=1.0\linewidth]{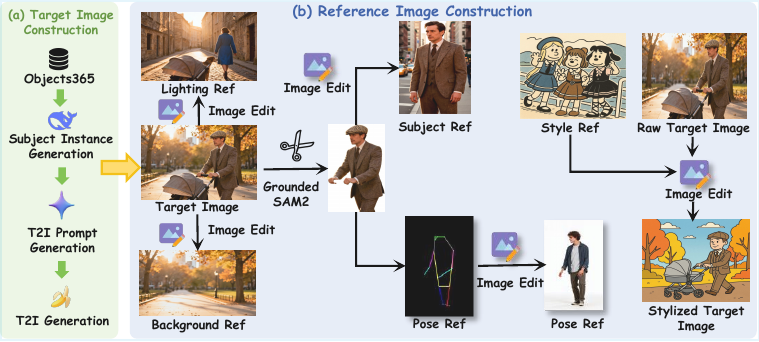}
\caption{\textbf{Data construction pipeline}. Our data construction pipeline is structured into two phases. \textbf{(a)} We obtain the target image using T2I generation. \textbf{(b)} We acquire the corresponding reference images of each type (except style) based on the target image, and we obtain the stylized target image by updating the original target image using an external style reference image from the OmniConsistency dataset~\cite{Song2025OmniConsistencyLS}.}
\label{fig:data pipeline}
\end{figure*}

\mypara{Benchmark Construction.}
OmniRef-Bench consists of both real-world and synthetic images. The real-world images are sourced from the LAION-5B dataset~\cite{schuhmann2022laion}. For synthetic image generation, we design an efficient multi-reference data generation pipeline to obtain high-quality data for both training and benchmarking. As illustrated in Fig.~\ref{fig:data pipeline}, inspired by UNO~\cite{wu2025less}, we obtain raw subject concepts from the Objects365 dataset~\cite{shao2019objects365}, and then utilize LLM~\cite{liu2024deepseek, team2023gemini} to generate subject instances and text-to-image (T2I) prompts sequentially. The generated prompts are then fed into the T2I generation models~\cite{seedream2025seedream, Nanobanana} to obtain the target image. 
Based on the generated target image, we then utilize a segmentation model~\cite{liu2023grounding, ravi2024sam2segmentimages} and a series of image editing models~\cite{wu2025qwenimagetechnicalreport} to produce each type of reference image. For more details, please refer to Appendix~\ref{subsec:data construct}.
\subsection{Evaluation Protocol}
\label{subsec:bench_eval}
As shown in Tab.~\ref{tab:bench_comparision}, OmniRef-Bench adopts both objective metrics and MLLM-based evaluation. The details are as follows.

\mypara{Objective Metrics.}
We assess the consistency between the generated image and the reference images for each of the five reference types. For each type, we either leverage established metrics~\cite{radford2021learning, oquab2023dinov2, somepalli2024measuring, huang2024vbench, chen2020monocular}, or propose carefully-designed custom metrics for dimensions where prior evaluation methods are lacking.
Please refer to Appendix~\ref{subsec:obj metrics} for more details.

\mypara{MLLM-based Evaluation.}
We use Gemini 3 Flash~\cite{team2023gemini} to conduct the evaluation. Compared with objective metrics, we remove the pose consistency evaluation based on our observation in Appendix~\ref{subsec:eval_prompt} that MLLM cannot accurately measure the fine-grained angles of human limbs. Meanwhile, we introduce aesthetic and instruction following dimensions to fully leverage the high-level semantic understanding ability of MLLM. Refer to Appendix~\ref{subsec:eval_prompt} for more details.

%% file: sections/method.tex
\section{Methods}
\label{sec:methods}
\begin{figure*}[t]
\centering
\includegraphics[width=\textwidth]{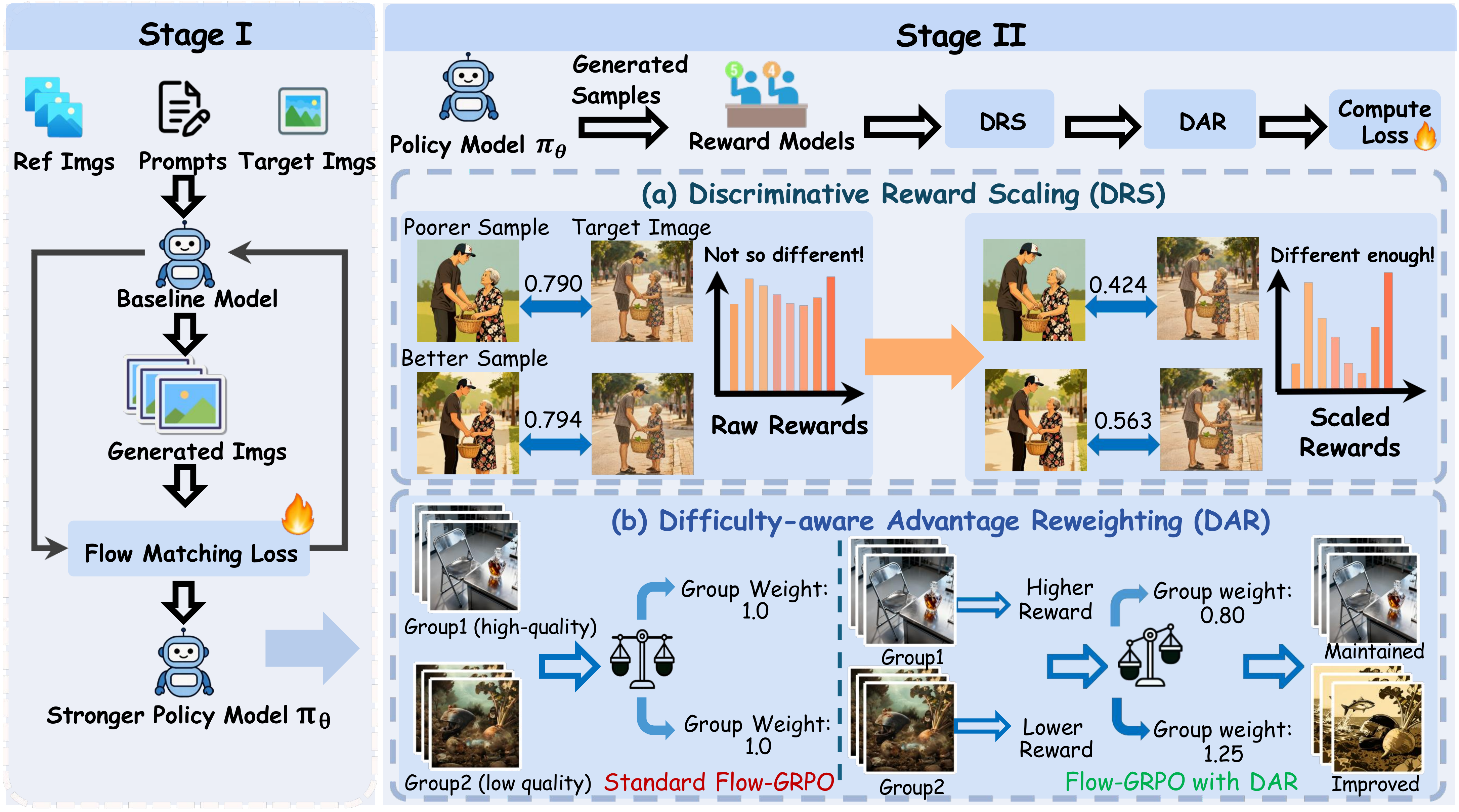}
\caption{\textbf{Framework of DyRef}. In Stage I, SFT equips the model with the basic capability to handle complex MRIG tasks. In Stage II, \textbf{(a)} DRS enlarges the reward differences across samples for better training. \textbf{(b)} DAR enhances the model’s focus on samples with a large number of mixed-type reference images.}
\label{fig:method_overview}
\end{figure*}
In Sec.~\ref{sec:datasets}, we introduce OmniRef-Bench to assess the complex multi-reference image generation (MRIG) task. Evaluation results (Tab.~\ref{tab:results_Omni-Ref}) indicate that mainstream open-source models perform poorly on this task. Further analysis (Fig.~\ref{fig:method_motivation}) shows that their performance declines rapidly as the number of reference images involving mixed reference types increases. To address this issue, we propose DyRef, a two-stage training framework. As shown in Fig.~\ref{fig:method_overview}, in Stage I, we employ SFT to equip the model with the basic capability to handle complex MRIG tasks. In Stage II, \textbf{(a)} DRS enlarges the reward differences across samples for better training, and \textbf{(b)} DAR enhances the model’s focus on samples with numerous mixed-type reference images. The combination of DAR and DRS enhances the model’s ability to learn from samples with a large number of diverse reference images, thereby improving performance on complex MRIG tasks.

\subsection{Stage I: Supervised Fine-Tuning}
In the first stage, we adopt Qwen-Image-Edit-2511~\cite{wu2025qwenimagetechnicalreport} as our training baseline, which employs a standard dual-stream MMDiT backbone~\cite{flux-2-2025}. 
This dual-stream mechanism enables the editing module to achieve a robust balance between maintaining high-level semantic consistency and preserving low-level visual fidelity across multiple inputs.
To equip the model with the basic capability to handle complex MRIG tasks, we construct approximately 14000 training samples using the data construction pipeline outlined in Fig.~\ref{fig:data pipeline}. We then fine-tune the model using LoRA~\cite{hu2022lora} with Flow Matching loss~\cite{lipman2022flow,liu2025flow,ping2026flowfactory}. 

Formally, let $z_0 \sim p_0(z)$ denote samples drawn from a standard Gaussian distribution, and let $z_1 \sim p_{\text{data}}(z)$ denote samples drawn from the target data distribution. $z_t$ represents the intermediate state along the transport trajectory that maps $z_0$ to $z_1$. Moreover, $v_\theta(z,t)$ denotes a time-dependent velocity field parameterized by a neural network, and $\theta$ denotes the learnable parameters of the model. The Flow Matching optimization objective is defined as:
\begin{equation}
\label{eq:flow_matching_loss}
\min_{v}\int_0^1 \mathbb{E}\!\left[\left\lVert (z_1 - z_0) - v_{\theta}(z_t,t)\right\rVert^2\right] \,\mathrm{d}t.
\end{equation}

\subsection{Stage II: Difficulty-aware Advantage Reweighting}
\label{subsec:drm-grpo}
As shown in Tab.~\ref{tab:results_Omni-Ref} and Tab.~\ref{tab:ablation}, although the model fine-tuned with SFT has acquired preliminary capabilities, a noticeable gap remains compared with the SOTA closed-source model~\cite{Nanobanana} on OmniRef-Bench, particularly in style-related tasks. 
Furthermore, we observe that the model performs poorly when processing complex samples containing numerous mixed-type reference images (Fig.~\ref{fig:method_motivation}). To address these limitations, we introduce a style-specific reward for targeted optimization of style-related tasks. In addition, we propose \textbf{D}ifficulty-aware \textbf{A}dvantage \textbf{R}eweighting (DAR) to enhance the model's focus on underperforming samples with a large number of diverse reference images. Notably, during the RL stage, we use data drawn from the same source as that used in the SFT stage, without introducing new datasets.

\mypara{Rewards.}
Inspired by the USO~\cite{wu2025uso} framework, we adopt CSD~\cite{somepalli2024measuring} as the reward model for style alignment to guide the model toward generating images that better match the style references. In addition, we employ CLIP~\cite{radford2021learning} or SigLIPv2~\cite{tschannen2025siglip} to measure the semantic consistency of training samples between generated images and their corresponding target images, thereby further reducing the distribution gap between generated outputs and the target data.

\mypara{Motivation.}
For MRIG tasks, the amount of visual information increases as the number of mixed-type reference images grows, making these samples more complex and requiring greater focus during training. However, standard Flow-GRPO~\cite{ping2026flowfactory,liu2024deepseek} lacks the ability to dynamically optimize for samples of varying complexity. As a result, 
training tends to be dominated by simpler samplers with fewer reference images, while more complex samples with numerous mixed-type reference images receive insufficient learning signals during training. 

In dense object detection, prior studies~\cite{lin2017focal} have addressed sample imbalance by designing loss functions that adjust each sample’s contribution to the overall loss, promoting targeted optimization. Furthermore, we observe that the reward scores of different samples also decrease as the number of mixed-type reference images increases. A detailed analysis is provided in Appendix~\ref{subsec:reward distribution}. Inspired by these observations, we propose DAR, an improved method based on Flow-GRPO. DAR dynamically adjusts the contribution of each sample group of Flow-GRPO to the total loss based on the average reward score of samples within the group, thereby encouraging the model to prioritize underperforming samples with more mixed-type reference images.

Specifically, consider a single optimization step with a training set of samples denoted by $\mathcal{S}$. 
We partition $\mathcal{S}$ into groups $\mathcal{G}$, 
where each group $g \in \mathcal{G}$ corresponds to multiple samples generated from the same prompt. 
Let $I_g \subset \{1, \dots, |\mathcal{S}|\}$ denote the index set of samples belonging to group $g$.
For each sample $i \in \mathcal{S}$, we obtain a scalar reward score $p_i \in [0, 1]$. 
We first compute the mean reward of each group $g$ as follows:
\begin{equation}
    p_g = \frac{1}{|I_g|} \sum_{i \in I_g} p_i.
\end{equation}
To emphasize groups with poor performance (i.e., those with lower average rewards), we construct a raw group weight based on the mean reward as follows:
\begin{equation}
    w_g^{\text{raw}} = \left(1 - p_g + \epsilon\right)^{\gamma},
\end{equation}
where $\epsilon > 0$ is a small constant for numerical stability and 
$\gamma > 0$ is a hyperparameter that controls the degree of reweighting. 
Then, the raw weight is clipped to the predefined range $[w_{\min}, w_{\max}]$ as follows:
\begin{equation}
    w_g = \mathrm{clip}\left(w_g^{\text{raw}},\, w_{\min},\, w_{\max}\right).
\end{equation}
To avoid unintended shifts in overall loss scale, we normalize group weights to a batch-wise mean of approximately 1.
Let $\mu$ denote the mean group weight within the current batch, and the final normalized group weight is defined as:
\begin{equation}
    \mu = \frac{1}{|\mathcal{G}|} \sum_{g \in \mathcal{G}} w_g, \quad \tilde{w}_g = 
    \mathrm{clip}\left(\frac{w_g}{\mu},\, w_{\min},\, w_{\max}\right).
\end{equation}
Then, each sample $i$ inherits the weight of its corresponding group $g(i)$:
\begin{equation}
    w_i = \tilde{w}_{g(i)}.
\end{equation}
Finally, the computed $w_i$ weights each sample’s contribution to the final objective. Following Flow-GRPO~\cite{liu2025flow}, $r_i(\theta)$ denotes the probability ratio between the current policy and the behavior policy for sample $i$, and $\hat{A}_i$ represents the estimated advantage of each sample. The weighted objective is defined as:
\begin{equation}
    L(\theta) = -\frac{1}{|\mathcal{S}|} \sum_{i \in \mathcal{S}} w_i \min \left( r_i(\theta) \hat{A}_i, \operatorname{clip}\left(r_i(\theta), 1-\epsilon, 1+\epsilon\right) \hat{A}_i \right).
\end{equation}


\mypara{Discriminative Reward Scaling.}
Although DAR is effective, we still face another challenge. As illustrated in Fig.~\ref{fig:method_overview}(a), we observe that the reward scores computed using CLIP or SigLIPv2 tend to be overly concentrated, and even when generated samples exhibit substantial differences in visual quality, their assigned reward scores remain highly similar. Consequently, when calculating intra-group advantages, insufficient numerical contrast between superior and inferior samples weakens the gradient signals for policy updates. This phenomenon may stem from intrinsic biases in the pre-trained evaluation models. Detailed quantitative analysis is provided in Appendix~\ref{subsec:reward distribution}.
To address this issue, we propose \textbf{D}iscriminative \textbf{R}eward \textbf{S}caling (DRS), which aims to enhance the discriminability of reward values through a transformation function. By enlarging reward differences between high-quality and low-quality samples, DRS preserves robust gradient signals for policy updates, improving overall model performance.

Specifically, at each optimization step of DAR, for every generated sample, let $F_{\text{gen}}$ denote the CLIP or SigLIPv2 feature of the generated image and $F_{\text{target}}$ denote the CLIP or SigLIPv2 feature of the target image, respectively. We first compute the cosine similarity $p = \cos(F_{\text{gen}}, F_{\text{target}})$ as the raw reward. Subsequently, we amplify the reward differences among samples within the same group using a function $F$:
\begin{equation}
\hat{p} = F(p, t),
\end{equation}
where $0<t<1$ is a hyperparameter that controls the distribution of rewards and $\hat{p}$ is a reward transformed by DRS.
A detailed analysis of the implementation of the function $F$ is provided in Appendix~\ref{subsec:analysis_F}.

%% file: sections/experiments.tex
\vspace{-2mm}
\section{Experiments}
\begin{table}[t]
\centering
\setlength{\tabcolsep}{2pt}  
\renewcommand{\arraystretch}{1.1}
\caption{Results of various methods on the OmniRef-Bench. \textbf{Bold} and \underline{underlined} indicate the best and second-best results, respectively. Gray background indicates implementations with our method. Bg, Aes, IF, and Avg are the abbreviations for background, aesthetic, instruction following, and average, respectively. Qwen-2511 and FLUX.2 [klein] denotes the Qwen-Image-Edit-2511 model and the FLUX.2 [klein] 9B Base model, respectively. All MLLM evaluations are conducted using Gemini 3 Flash.}
\resizebox{\textwidth}{!}{%
\begin{tabular}{lccccccccccccc}
\toprule
\multirow{2.5}{*}{Methods} & \multicolumn{6}{c}{Objective Metrics}                                                           & \multicolumn{7}{c}{MLLM Evaluation}                                                                            \\ \cmidrule(lr){2-7} \cmidrule(lr){8-14} 
                               & Subject       & Style         & Bg            & Light         & Pose          & Avg           & Subject       & Bg            & Style         & Light         & Aes           & IF            & Avg           \\ \midrule
\multicolumn{14}{c}{Closed-Source Models} \\ \midrule
Seedream4.5~\cite{seedream2025seedream}                    & 0.65 & 0.34    & 0.89 & 0.74          & 0.74    & 0.67          & 8.77 & 8.92 & 6.27          & 8.78 & 8.08          & 7.97          & 8.13          \\
Nano Banana Pro~\cite{Nanobanana}                & 0.64    & 0.46 & 0.89 & 0.77 & 0.72          & 0.70    & 8.50    & 8.61    & 8.37    & 8.75    & 8.37 & 8.39 & 8.50 \\ \midrule
\multicolumn{14}{c}{Open-Source Models} \\ \midrule
OmniGen2~\cite{wu2025omnigen2}                       & 0.59          & 0.11          & 0.85          & 0.65          & 0.65          & 0.57          & 4.47          & 2.31          & 0.37          & 4.68          & 6.73          & 4.16          & 3.79          \\
DreamOmni2~\cite{xia2025dreamomni2}                     & 0.57          & 0.28          & 0.85          & 0.71          & 0.67          & 0.62          & 6.75          & 3.43          & 3.45          & 6.07          & 6.65          & 4.71          & 5.18          \\
BAGEL~\cite{deng2025bagel}                          & 0.58          & 0.15          & 0.85          & 0.67          & 0.69          & 0.59          & 6.21          & 3.66          & 1.13          & 6.31          & 5.98          & 4.54          & 4.64          \\
\midrule
Qwen-2511~\cite{wu2025qwenimagetechnicalreport}           & 0.55          & 0.15          & 0.84          & 0.71          & 0.71          & 0.59          & 6.50          & 4.35          & 1.61          & 6.81          & 6.09          & 4.46          & 4.97         \\ \rowcolor{gray!15}
\hspace{0.5em}+ Ours          & \underline{0.62}          & \underline{0.46} & \underline{0.87}    & \underline{0.76}   & \textbf{0.83} & \textbf{0.71} & \textbf{8.34}          & \textbf{8.51}          & \textbf{8.67} & \textbf{8.54}          & \textbf{8.15}    & \textbf{8.08}    & \textbf{8.38}    \\ 
FLUX.2 [klein]~\cite{flux-2-2025}           & \textbf{0.63}          & 0.21          & \underline{0.87}    & 0.69          & 0.74    & \underline{0.63}          & \underline{8.25}          & 6.32          & 2.18          & 7.69          & 7.95          & 6.53          & 6.49          \\  \rowcolor{gray!15}
\hspace{0.5em}+ Ours           & \textbf{0.63}          & \textbf{0.48}          & \textbf{0.90}    & \textbf{0.78}          & \underline{0.76}    & \textbf{0.71}          & 8.07          & \underline{8.12}          & \underline{8.06}          & \underline{8.35}          & \underline{8.11}          & \underline{7.48}          & \underline{8.03}          \\
\bottomrule
\end{tabular}%
}
\label{tab:results_Omni-Ref}
\end{table}
In this section, we present the main results of our proposed method in Sec.~\ref{exp:main results} and analyze our design choices via ablation studies in Sec.~\ref{exp:ablation}. Details about our training and experiment settings can be found in Appendix~\ref{sec:training details} and~\ref{sec:eval details}, respectively. Detailed qualitative analysis and failure cases are provided in Appendix~\ref{sec:case study}.

\subsection{Main Results}
\label{exp:main results}
First, we conduct extensive experiments on our proposed OmniRef-Bench and two other single-image editing benchmarks~\cite{peng2024dreambench, ye2025imgedit}. To verify the generalizability of our method on models of different parameter scales, we select Qwen-Image-Edit-2511 (Qwen-2511, 20B) and FLUX.2 [klein] 9B Base (FLUX.2 [klein]) as our base models and implement our method on them. We compare the results of our method with SOTA closed-source models~\cite{Nanobanana, seedream2025seedream} and mainstream open-source models~\cite{wu2025omnigen2, xia2025dreamomni2, deng2025bagel}. Subsequently, we analyze the alignment of our evaluation results with human preference through qualitative analysis and a user study. More experimental results can be found in Appendix~\ref{sec:additional results}.

\mypara{Results on OmniRef-Bench.}
To verify the effectiveness of our method on the complex multi-reference generation task, we conduct experiments on the proposed OmniRef-Bench. As shown in Tab.~\ref{tab:results_Omni-Ref}, Qwen-2511 with our method significantly outperforms all open-source models by at least 12\% (+0.08 compared with FLUX.2 [klein]) and 29\% (+1.89 compared with FLUX.2 [klein]) in objective metrics and MLLM evaluation, respectively, with a particular enhancement in style and pose-related metrics. Meanwhile, it also surpasses the two closed-source models in objective metrics and remains comparable with Nano Banana Pro in MLLM evaluation. These results demonstrate that our method can handle diverse reference combinations with varying reference counts and achieve competitive performance with closed-source models.

\mypara{Results on Single-Image Editing Benchmarks.}
\begin{table}[t]
\centering
\setlength{\tabcolsep}{4pt}  
\renewcommand{\arraystretch}{1.1}
\caption{Performance evaluation on the ImgEdit Benchmark. All results are evaluated by GPT-4o following the default settings. Results on closed-source models are sourced from UniRef-Image-Edit~\cite{wei2026uniref}.}
\begin{tabular}{lcccccccccc}
\toprule
Methods         & Extract       & Add           & Style         & Bg            & Adjust        & Replace       & Remove        & Hybrid        & Action        & \multicolumn{1}{l}{Overall} \\ \midrule
\multicolumn{11}{c}{Closed-Source Models} \\ \midrule
Seedream4.0~\cite{seedream2025seedream}           & 2.39          & 4.52    & 4.76    & 4.30    & 4.41 & 4.56 & 4.44 & 3.33          & 4.36          & 4.18                        \\
GPT Image 1 {[}High{]} & 2.90          & 4.61 & 4.93 & 4.57 & 4.33    & 4.35          & 3.66          & 3.96    & 4.89 & 4.20                  \\ \midrule
\multicolumn{11}{c}{Open-Source Models} \\ \midrule
OmniGen2~\cite{wu2025omnigen2}              & 2.43          & 3.72          & 4.63          & 3.72          & 3.11          & 3.75          & 2.62          & 2.60          & 4.52          & 3.46                        \\
DreamOmni2~\cite{xia2025dreamomni2}            & 2.95          & 3.83          & 4.43          & 3.90          & 3.53          & 4.27          & 3.09          & 2.91          & 4.50          & 3.71                        \\
BAGEL~\cite{deng2025bagel}                 & 1.77          & 3.61          & 4.20          & 3.33          & 3.40          & 3.76          & 2.59          & 2.59          & 4.32          & 3.29                        \\
\midrule
Qwen-2511~\cite{wu2025qwenimagetechnicalreport}  & \textbf{4.23} & \underline{4.40}          & 4.61          & \textbf{4.23}          & 3.91          & 4.21          & 3.76          & 2.72          & 4.52          & 4.07                        \\ \rowcolor{gray!15}
\hspace{0.5em}+ Ours & \underline{4.07}    & 4.35          & 4.66          & \underline{4.17}          & \textbf{4.10}          & 4.46          & \underline{4.12}    & \textbf{4.21} & \underline{4.72}    & \textbf{4.32}               \\ 
FLUX.2 [klein]~\cite{flux-2-2025}  & 2.89          & 4.36          & \textbf{4.72}          & 3.99          & \underline{4.02}          & \underline{4.47}    & 3.95          & 3.50          & \underline{4.72}    & 4.07                        \\ 
\rowcolor{gray!15}
\hspace{0.5em}+ Ours  & 2.81          & \textbf{4.43}          & \underline{4.69}          & 4.08          & 3.84          & \textbf{4.48}    & \textbf{4.27}          & \underline{3.65}          & \textbf{4.77}    & \underline{4.11}                        \\ 
\bottomrule
\end{tabular}
\label{tab:results_imgedit}
\end{table}
\begin{table}[t]
\centering
\setlength{\tabcolsep}{4pt}  
\renewcommand{\arraystretch}{1.1}
\caption{Performance on the DreamBench++. Photo., Style., Imag. are the abbreviations for Photorealistic, Style Transfer, and Imaginative, respectively. All results are evaluated by GPT-4o.}
\begin{tabular}{lcccccccccc}
\toprule
\multirow{2.5}{*}{\textbf{Methods}} & \multicolumn{5}{c}{\textbf{Concept Preservation (CP)}} & \multicolumn{4}{c}{\textbf{Prompt Following (PF)}}  & \multirow{2.5}{*}{\textbf{CP*PF}} \\ \cmidrule(lr){2-6} \cmidrule(lr){7-10}
                                 & Animal     & Human     & Object    & Style    & Overall          & Photo. & Style. & Imag. & Overall       &                                 \\ \midrule
OmniGen2~\cite{wu2025omnigen2}                         & 0.54       & 0.41      & 0.55      & 0.49     & 0.52             & \underline{0.95}           & 0.93           & 0.87        & 0.92          & 0.48                            \\
DreamOmni2~\cite{xia2025dreamomni2}                       & 0.64       & \underline{0.59}      & \underline{0.67}      & 0.53     & 0.63      & 0.81           & 0.80           & 0.7         & 0.78          & 0.49                            \\
BAGEL~\cite{deng2025bagel}                            & 0.45       & 0.23      & 0.38      & 0.46     & 0.39             & \textbf{0.97}           & \textbf{0.97}           & 0.91        & \underline{0.95}    & 0.37                            \\
\midrule
Qwen-2511                         & 0.58       & 0.55      & 0.59      & 0.54     & 0.58             & 0.90           & \underline{0.94}           & \underline{0.93}        & 0.92          & 0.53                            \\ \rowcolor{gray!15}
\hspace{0.5em}+ Ours                           & \textbf{0.68}       & \textbf{0.61}      & \textbf{0.69}      & \textbf{0.70}     & \textbf{0.68}    & 0.90           & \underline{0.94}           & \underline{0.93}        & 0.92          & \textbf{0.62}                   \\ 
FLUX.2 [klein]~\cite{flux-2-2025}                     & 0.59       & 0.53      & 0.61      & 0.54     & 0.58             & \textbf{0.97}           & \textbf{0.97}           & \textbf{0.94}        & \textbf{0.97} & 0.56                      \\ 
\rowcolor{gray!15}
\hspace{0.5em}+ Ours                   & \underline{0.65}       & 0.58      & 0.66      & \underline{0.61}     & \underline{0.64}             & 0.94           & \textbf{0.97}           & \underline{0.93}        & \underline{0.95} & \underline{0.61}                      \\ 
\bottomrule
\end{tabular}
\label{tab:results_dreambench}
\end{table}
\begin{table}[t]
\centering
\setlength{\tabcolsep}{6pt}  
\renewcommand{\arraystretch}{1.1}
\caption{User Study. (1-10) and (1-4) denote the range of the scores. Nano denotes Nano Banana Pro. PLCC, SROCC, and KROCC denote Pearson, Spearman, and Kendall correlation coefficients, respectively.}
\begin{tabular}{lccccccc}
\toprule
\multirow{2.5}{*}{\textbf{Evaluation Setup}} & \multicolumn{4}{c}{\textbf{Evaluation Results}}                                       & \multicolumn{3}{c}{\textbf{Correlation Analysis}}                               \\ \cmidrule(lr){2-5} \cmidrule(lr){6-8} 
                                  & Ours & Nano & Seedream4.5 & Qwen-2511 & PLCC             & SROCC        & KROCC           \\ \midrule
Gemini3 Flash (1-10)              & \underline{8.38}           & \textbf{8.50}   & 8.13         & 4.97                 & \multirow{2}{*}{0.902} & \multirow{2}{*}{0.800} & \multirow{2}{*}{0.667} \\
Human (1-4)                       & \textbf{3.42}        & \underline{3.09}      & 2.26         & 1.22                 &                        &                      &                        \\ \bottomrule
\end{tabular}
\label{tab:user_study}
\end{table}
To ensure our method enhances the base model's multi-reference generation ability without compromising its basic image editing capability, we conduct experiments on two single-image editing benchmarks, ImgEdit~\cite{ye2025imgedit} and DreamBench++~\cite{peng2024dreambench}. The experimental results are reported in Tab.~\ref{tab:results_imgedit} and Tab.~\ref{tab:results_dreambench}, respectively. Results on both benchmarks demonstrate that our method achieves superior performance over both the base models and other open-source models. Specifically, on ImgEdit, Qwen-2511 with our method not only significantly surpasses all open-source models with the highest score of 4.32, but also outperforms closed-source models such as  Seedream4.0~\cite{seedream2025seedream} and GPT Image 1 [high]. On DreamBench++, Qwen-2511 with our method achieves the highest overall score of 0.62. For the results of the objective metrics on DreamBench++, please refer to Appendix~\ref{subsec:dreambench++_result}. These results indicate that our method not only improves the base model's multi-reference generation ability, but also enhances its single-image editing ability.

\mypara{Generalizability of Our Method.}
As shown in Tab.~\ref{tab:results_Omni-Ref}, Tab.~\ref{tab:results_imgedit}, and Tab.~\ref{tab:results_dreambench}, integrating our method with FLUX.2 [klein] not only significantly outperforms the base model on the multi-reference generation task (8.03 vs. 6.49), but also enhances its single-image editing capabilities (e.g., improving from 4.07 to 4.11 on ImgEdit and 0.56 to 0.61 on DreamBench++). Furthermore, despite having a smaller parameter scale (9B) compared to Qwen-2511 (20B), our approach achieves competitive performance, underscoring its effectiveness and generalizability across models of varying scales.


\mypara{User Study.}
To evaluate the consistency between MLLM-based scoring and human preferences, we conduct a user study. We invite 30 experts to rank 20 randomly sampled examples. The evaluation considers text fidelity, reference similarity, composition quality, visual appeal, and task completion, with detailed settings provided in Appendix~\ref{sec:eval details}. As shown in Tab.~\ref{tab:user_study}, ours and Nano Banana Pro rank in the top two, while Seedream 4.5 and Qwen-Image-Edit-2511 rank third and fourth, respectively. Furthermore, the Pearson correlation coefficient between the two is 0.9, indicating strong linear agreement, while the Spearman correlation coefficient is 0.8, suggesting high consistency in ranking. These results show that LLM-based scoring aligns well with human preferences.

\subsection{Ablation}
\label{exp:ablation}
\begin{table}[t]
\centering
\setlength{\tabcolsep}{4pt}  
\renewcommand{\arraystretch}{1.1}
\caption{Ablation study on the OmniRef-Bench. }
\begin{tabular}{lccccccccccccc}
\toprule
\multirow{2}{*}{Settings} & \multicolumn{6}{c}{Objective Metrics}                                                         & \multicolumn{7}{c}{MLLM Evaluation}                                                                            \\ \cmidrule(lr){2-7} \cmidrule(lr){8-14}
                        & Subject       & Style         & Bg            & Light         & Pose          & Avg           & Subject       & Bg            & Style         & Light         & Aes           & IF            & Avg           \\ \midrule
SFT                     & 0.62          & 0.38          & \textbf{0.88} & 0.72          & 0.82          & 0.68          & 8.20          & 8.04          & 7.57          & 8.13          & 7.75          & 7.72          & 7.90          \\ \midrule \rowcolor{gray!15}
Ours                    & 0.62          & \textbf{0.46} & \underline{0.87}    & \textbf{0.76} & \underline{0.83}    & \textbf{0.71} & 8.34          & \underline{8.51}    & \textbf{8.67} & 8.54          & \textbf{8.14} & \textbf{8.08} & \textbf{8.38} \\
\hspace{0.5em}w/o CSD               & \textbf{0.64} & 0.36          & \textbf{0.88} & 0.72          & \underline{0.83}    & \underline{0.69}    & 8.52          & \textbf{8.79} & 6.33          & 8.42          & \underline{8.05}    & 7.91          & 8.00          \\
\hspace{0.5em}w/o DAR              & \underline{0.63}    & 0.39          & \textbf{0.88} & 0.74          & 0.81          & \underline{0.69}    & \textbf{8.54} & 8.39          & 6.98          & \textbf{8.58} & 7.89          & 7.82          & 8.03          \\
\hspace{0.5em}w/o DRS               & \underline{0.63}    & 0.45          & \textbf{0.88} & \underline{0.75}    & \textbf{0.84} & \textbf{0.71} & \underline{8.53}    & 8.41          & \underline{8.12}    & \underline{8.55}    & 8.03          & \underline{8.01}    & \underline{8.28}\\ \bottomrule
\end{tabular}
\label{tab:ablation}
\end{table}
We conduct ablation studies on Qwen-Image-Edit-2511 to verify the effectiveness of three of our key training components. The experimental results are reported in Tab.~\ref{tab:ablation}, where SFT denotes the results after the first stage of supervised fine-tuning. Comparisons are made with SFT as the baseline. More ablation studies on hyperparameter sensitivity analysis are provided in Appendix~\ref{subsec:hyper_analysis}.

\mypara{Effect of CSD.}
CSD score is added to our reward to improve the stylization ability of the model. As shown in the third row, removing the CSD reward substantially impairs the model's performance in style-related aspects. Specifically, the objective style metric drops from 0.46 to 0.36, and the style metric in MLLM evaluation drops from 8.67 to 6.33, which confirms that the CSD reward is an effective way to improve the model's stylization ability.

\mypara{Effect of DAR.}
Removing DAR degrades the training process into standard GRPO, which has limited discrimination between simple and difficult samples. As shown in the fourth row, both objective metrics and MLLM evaluation results suffer a noticeable decline (0.71 to 0.69 and 8.38 to 8.03, respectively). Notably, even with CSD reward, the style-related score is low compared with our complete training settings (second row), emphasizing DAR's important role to tackle difficult combinations of reference types. 

\mypara{Effect of DRS.}
As stated in Sec.~\ref{subsec:drm-grpo}, DRS is introduced to magnify the inherent insufficient numerical contrast of objective reward scores. The last row of Tab.~\ref{tab:ablation} shows that while objective metrics remain relatively stable after removing DRS, several MLLM evaluation metrics experience a nontrivial decline. For example, the Style score drops from 8.67 to 8.12, Aes score drops from 8.14 to 8.03. This verifies that DRS can improve the model's performance in high-level semantic aspects, which cannot be effectively captured by low-level objective scores.

\begin{figure*}[t]
\centering
\includegraphics[width=0.95\linewidth]{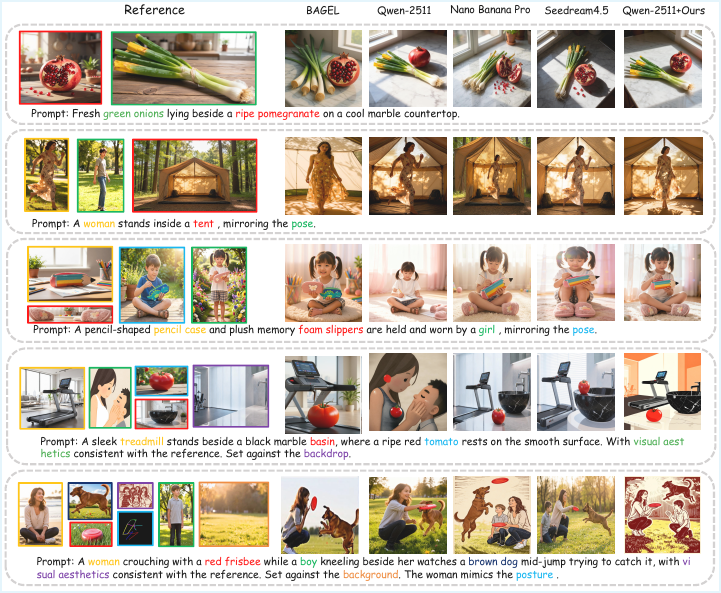}
\caption{Qualitative comparison with different models or methods on OmniRef-Bench. Each reference image is framed with a distinct color. Its corresponding phrase in the text prompt is also displayed in the same color.}
\label{fig:qualitative_analysis}
\end{figure*}

%% file: sections/conclusion.tex
\section{Conclusion}

We propose DyRef to advance multi-reference image generation (MRIG) by addressing the critical performance degradation observed as reference complexity and counts increase. By incorporating the Difficulty-aware Advantage Reweighting and Discriminative Reward Scaling, our approach effectively balances optimization across varying reference scales to mitigate performance variance. Experiments demonstrate that DyRef substantially improves performance on complex MRIG and single-image editing tasks, demonstrating the effectiveness and generalization capability of our approach.

%% file: sections/appendix.tex
\newpage

\appendix

\section*{\hspace{-4mm} \centering Appendix}
\vspace{3mm}
\section*{Overview}
This material provides supplementary details to the main
paper, including the following sections:
\setcounter{section}{0}
\renewcommand{\thesection}{\Alph{section}}  
\renewcommand{\theHsection}{appendix.\arabic{section}} 
    




{
    \setlength{\parskip}{3pt}
    \vspace{0.5em}

    \apptocA{sec:related works}{A \quad Related Work}

    \apptocA{sec:benchmark details}{B \quad Dataset and Benchmark Details}
    \apptocB{subsec:statistic_bench}{B.1 \quad Statistics of OmniRef-Bench}
    \apptocB{subsec:data construct}{B.2 \quad Details of Data Construction Pipeline}
    \apptocB{subsec:obj metrics}{B.3 \quad Details of Objective Metrics}
    \apptocB{subsec:eval_prompt}{B.4 \quad Prompts for MLLM-based Evaluation}
    \apptocB{subsec:comparison of obj and mllm}{B.5 \quad Limitations of Objective Metrics and MLLM Evaluation}

    \apptocA{sec:additional results}{C \quad Additional Experiments and Analysis}
    \apptocB{subsec:dreambench++_result}{C.1 \quad More Results on DreamBench++}
    \apptocB{subsec:hyper_analysis}{C.2 \quad Hyperparameter Sensitivity Analysis}
    \apptocB{subsec:reward distribution}{C.3 \quad Motivation of Difficulty-aware Advantage Reweighting}
    \apptocB{subsec:analysis_F}{C.4 \quad Detailed Analysis of $F$}
    \apptocB{subsec:comparision_clip_siglipv2}{C.5 \quad Comparison Between CLIP and SigLIPv2}
    \apptocB{subsec:omnicontext_multibanana_results}{C.6 \quad Additional Results on OmniContext and MultiBanana}
    \apptocB{subsec:more_ref_user}{C.7 \quad Generalization to More Reference Types and User Study}

    \apptocA{sec:training details}{D \quad Training Details}

    \apptocA{sec:eval details}{E \quad Evaluation Details}
    \apptocB{subsec:bench details}{E.1 \quad Benchmarks and Baselines}
    \apptocB{subsec:baselies}{E.2 \quad Backbones and Baselines}
    \apptocB{subsec:user study}{E.3 \quad Details of User Study}
    \apptocB{subsec:reproducibility}{E.4 \quad Reproducibility}

    \apptocA{sec:case study}{F \quad Case Study}
}

\input{sections/relatedwork}
\input{sections/appdix_dataset_details}
\input{sections/appdix_additional_exp}
\input{sections/appdix_train_details}
\input{sections/appdix_eval_details}
\input{sections/appdix_case_study}

%% file: sections/relatedwork.tex
\section{Related Work}
\label{sec:related works}
\mypara{Single-Image Editing.}
In recent years, personalized image generation~\cite{wu2025omnigen2,wu2025less,cheng2025umo, she2025mosaic} has advanced rapidly and demonstrated significant practical potential. Among various personalization paradigms, single-image editing has evolved from early style transfer tasks to more advanced semantic editing paradigms guided by user-provided textual instructions.
Early approaches are primarily based on inversion frameworks, such as Null-Text Inversion~\cite{mokady2023null}. These methods reconstruct the original image through the optimization of latent representations. They typically require extensive iterative optimization, leading to high computational costs and limited generalization performance. Subsequent research introduced more explicit spatial guidance mechanisms, including ControlNet~\cite{zhang2023adding} and IP-Adapter~\cite{ye2023ip}. By explicitly modeling structural constraints or incorporating image feature guidance, these methods~\cite{zhang2025context,liu2025step1x} enable more precise alignment between generated outputs and the corresponding user prompts.

Furthermore, a series of recent studies focuses on integrating multimodal large language models (MLLMs) into the personalized image generation pipeline, thereby substantially enhancing the capacity of image generation models to understand complex multimodal user instructions and align their outputs with human preferences. Representative works include BAGEL~\cite{deng2025emerging}, OmniGen2~\cite{wu2025omnigen2}, and Qwen-Image~\cite{wu2025qwenimagetechnicalreport}.
These models provide a unified and robust foundation for single-turn visual editing, thereby facilitating the practical deployment of single-image editing applications.

\mypara{Multi-Reference Image Generation.}
Although single-image editing models have achieved remarkable results, relying on purely textual instructions is often insufficient to accurately capture complex user requirements in artistic scenarios. Consequently, multi-reference image generation (MRIG)~\cite{wu2025less, wu2025uso, chen2025unireal} has gradually become a prominent research direction due to its ability to integrate diverse types of reference images based on user instructions.
Existing research on MRIG primarily focus on multi-subject image generation tasks~\cite{jin2025focusdpo, cheng2025umo,xu2025contextgen,mou2025dreamo}, such as XVerse~\cite{chen2025xverse}, MOSAIC~\cite{she2025mosaic}, and PSR~\cite{wang2025psr}. These methods typically aim to preserve the identity of each subject in the generated output, ensuring consistency with their corresponding reference images. 

However, these approaches are designed to preserve concrete concepts, such as the identity of a person or object. They are less effective in multi-reference scenarios that require the integration of both concrete and abstract attributes, such as style and atmospheric effects. Consequently, their applicability remains limited in complex real-world scenarios.
Recently, several studies have begun exploring MRIG that integrates mixture reference types~\cite{xia2025dreamomni2,xia2025dreamomni3}. They attempt to enhance multi-source information fusion by incorporating memory modules~\cite{song20253sgen}, reinforcement learning strategies~\cite{wei2026uniref,wang2025psr}, and agent-based mechanisms~\cite{oshima2025multibanana}. 
However, they also face challenges. On the one hand, the introduction of additional modules substantially increases both training and inference costs. On the other hand, the combinations of reference types considered are limited, typically involving only two types, which restricts their ability to address complex and diverse requirements in real-world scenarios.

\mypara{Reinforcement Learning for Image Generation.}
Reinforcement Learning (RL)~\cite{ouyang2022training,liu2024deepseek} has become one of the standard approaches for aligning large language models (LLMs) with human preferences during post-training. In recent years, as personalized image generation models have achieved significant advancements, the introduction of RL into image generation has emerged as a new research direction. Flow-GRPO~\cite{liu2025flow} marks an important milestone in this research direction, which first applies GRPO to text-to-image generation by optimizing flow-matching models with human preference signals, thereby improving the overall quality of generated images.
Building upon this foundation, Skywork UniPic 2.0~\cite{wei2025skywork} further advances the application of RL to visual generation tasks. It represents the first RL system specifically designed for single-image editing. By designing reward functions that encourage accurate comprehension and execution of editing instructions, the method greatly improves instruction adherence in the generated images.

Recent studies, such as PSR~\cite{wang2025psr} and UniRef-Image-Edit~\cite{wei2026uniref}, attempt to extend the RL framework to MRIG tasks, enabling models to leverage diverse reference images for image generation or editing. 
However, these approaches typically consider simple multi-reference settings, which involve a limited number of mixed-type reference images. This is insufficient for complex MRIG scenarios involving a large number of reference images and diverse reference types. Therefore, we introduce OmniRef-Bench to evaluate model performance on complex MRIG tasks and propose DyRef to improve performance on this task.

%% file: sections/appdix_dataset_details.tex
\section{Dataset and Benchmark Details}
\label{sec:benchmark details}

\subsection{Statistics of OmniRef-Bench}
\label{subsec:statistic_bench}
\begin{figure*}[tbp]
\centering
\includegraphics[width=\linewidth]{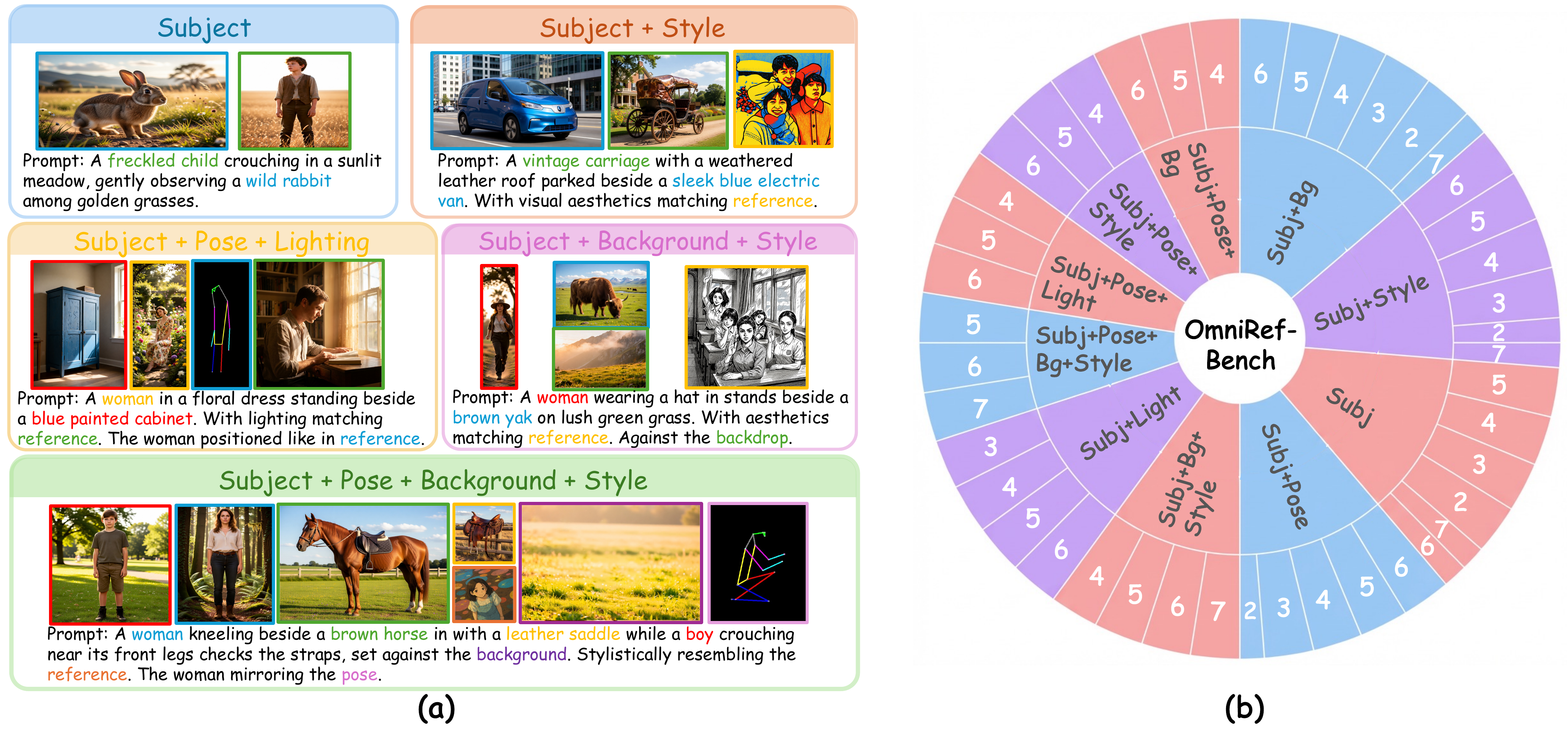}
\caption{\textbf{Overview of OmniRef-Bench}. \textbf{(a)} Some illustrative examples from OmniRef-Bench. \textbf{(b)} Sunburst chart showing the dataset distribution. Subj, Bg, and Light are the abbreviations for Subject, Background, and Lighting, respectively. The inner rim categorizes samples by reference type combinations. The outer rim further partitions these categories by the number of reference images. Notably, while the "Subj" task starts with two references, complex tasks with more reference types naturally require a higher lower bound of reference images to provide sufficient visual information.}
\label{fig:data statistics}
\end{figure*}

As illustrated in Fig.~\ref{fig:data statistics} (b), OmniRef-Bench encompasses a broad spectrum of complex reference type combinations. It is built upon five common reference types: Subject, Background (Bg), Style, Lighting (Light), and Pose. The benchmark comprises 395 high-quality, expert-annotated samples and spans 10 distinct combination categories, ranging from single-type (e.g., Subject) to highly complex multi-type scenarios (e.g., Subject + Pose + Bg + Style). These categories are further stratified by the number of reference images, which scales with the number of subject references. Notably, the benchmark maintains a balanced distribution across all combination categories and their respective sub-categories.

Regarding data composition, OmniRef-Bench integrates both synthetic and real-world images. The majority of the synthetic data is produced via our data construction pipeline (Sec.~\ref{subsec:data construct}). To ensure broad coverage, we supplement this with curated real-world images from LAION-5B~\cite{schuhmann2022laion} and additional high-quality samples, taking advantage of the selection results of MultiBanana~\cite{oshima2025multibanana}.

\begin{figure*}[tbp]
\centering
\includegraphics[width=\linewidth]{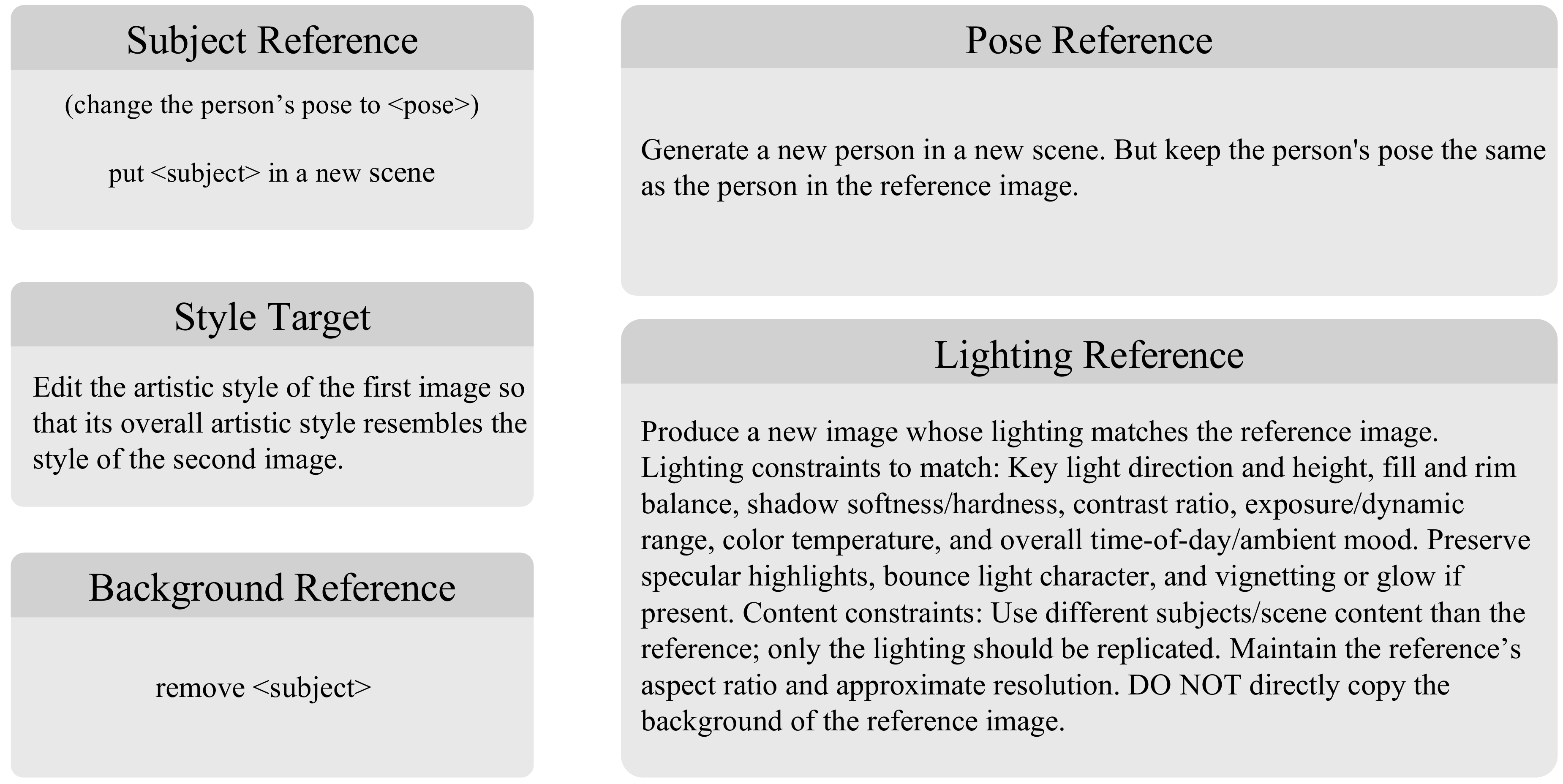}
\caption{The prompts for reference image (or target image) generation of each type. The placeholders <subject> and <pose> are replaced with specific entity names and posture descriptions during generation. Note that pose-related instruction in subject reference generation is exclusively applied when the subject is identified as a human.}
\label{fig:ref-gen prompts}
\end{figure*}

\subsection{Details of Data Construction Pipeline}
\label{subsec:data construct}
Our data construction pipeline is structured into two phases. Initially, we leverage text-to-image (T2I) models to synthesize the target images (ground-truth images for SFT training). Subsequently, we use a suite of tools to produce corresponding reference images for each of the five reference types. After obtaining the target images and the reference images, we conduct data filtering to ensure the quality of the generated images.

\mypara{T2I Generation for Target Image.}
Inspired by UNO~\cite{wu2025less}, we first obtain raw subject concepts (e.g., Person, Hat) from the Objects365 dataset~\cite{shao2019objects365}. Based on the subject concepts, we then utilize DeepSeek-v3~\cite{liu2024deepseek} to generate concrete subject instances (e.g., a man wearing a suit, a newsboy cap) as a way to diversify the subject categories. Next, we instruct Gemini 3 Flash~\cite{team2023gemini} to synthesize these instances into coherent scene descriptions, which serve as the T2I-prompts. Finally, we utilize SOTA image generation models (Seedream4.5~\cite{seedream2025seedream} or Nano Banana Pro~\cite{Nanobanana}) to complete the T2I generation and obtain the target images. The prompts for the above steps are shown in Fig.~\ref{fig:subject instance generation prompt} and Fig.~\ref{fig:prompt for t2i prompt generation}, respectively.
\begin{figure*}[tbp]
\centering
\includegraphics[width=\linewidth]{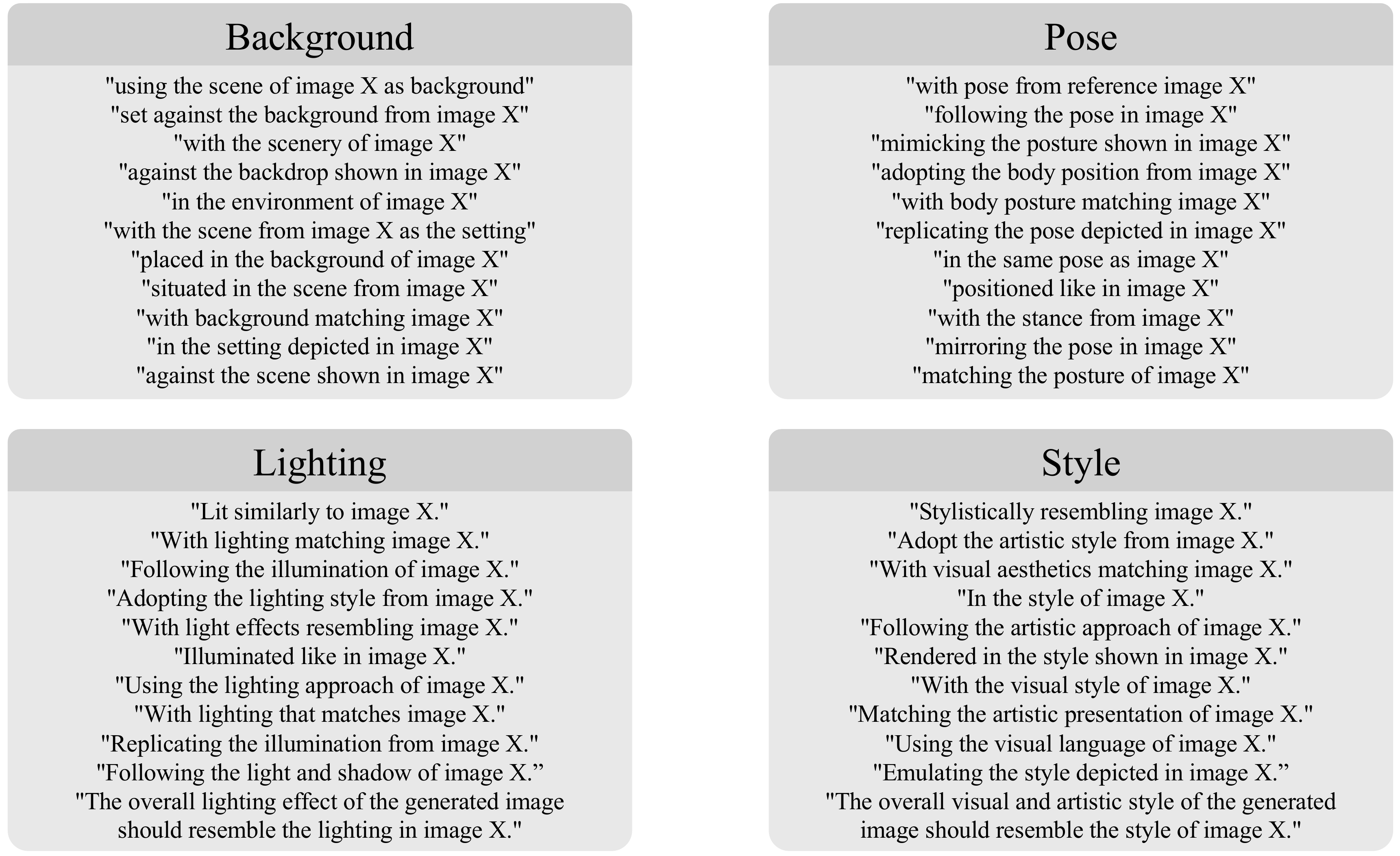}
\caption{Examples of diverse reference pointers for background, pose, lighting, and style. The placeholder "X" denotes the index of the reference image. We employ multiple linguistic variants for each reference type to enhance prompt diversity and robustness.}
\label{fig:reference pointer}
\end{figure*}

\mypara{Reference Image Construction.}
With the target images obtained, we first employ Grounded-SAM2~\cite{liu2023grounding,ravi2024sam2segmentimages} to segment the subjects in the target images according to the subject instance names. In this process, we obtain both the segmented subjects and their corresponding masks. Then we perform customized reference image generation for each reference type, as detailed below.

\textbf{Subject Reference.} We use the segmented subjects as raw subjects, and utilize Qwen-Image-Edit-2509~\cite{wu2025qwenimagetechnicalreport} to modify the subjects' background and pose (if necessary), as a way of data augmentation and to avoid the copy-paste effect during training. The modified subjects are used as the subject references.

\textbf{Background Reference.} We use Qwen-Image-Edit-2509 to perform subject removal on the target images according to the subject instances. The edited images with the specified subjects removed serve as the background references.

\textbf{Style Reference.} We first randomly assign an external style reference image from the OmniConsistency dataset~\cite{Song2025OmniConsistencyLS} for each target image. We then apply either the open-sourced stylization model USO~\cite{wu2025uso} or SOTA closed-source models~\cite{seedream2025seedream,Nanobanana} to stylize the original target images. These stylized target images then function as the ground-truth targets for training pairs involving style references.

\textbf{Lighting Reference.} Using the target images as reference, we leverage Seedream4.5 to generate new images that match the illumination effects of the target images. The generated images are used as the lighting references.

\textbf{Pose Reference.} We first utilize a pose estimation model (AlphaPose~\cite{alphapose}) to obtain the pose skeletons from the segmented human subjects. The pose skeletons can be directly used as the pose references. We also use some of the segmented human subjects as prompts for Seedream4.5 to generate new human figures, maintaining the same posture. Both the raw skeletons and the newly generated images are employed as pose references to ensure diversity.

Prompts for reference image (or target image) generation of each type are displayed in Fig.~\ref{fig:ref-gen prompts}.
\begin{figure*}[tbp]
\centering
\includegraphics[width=\linewidth]{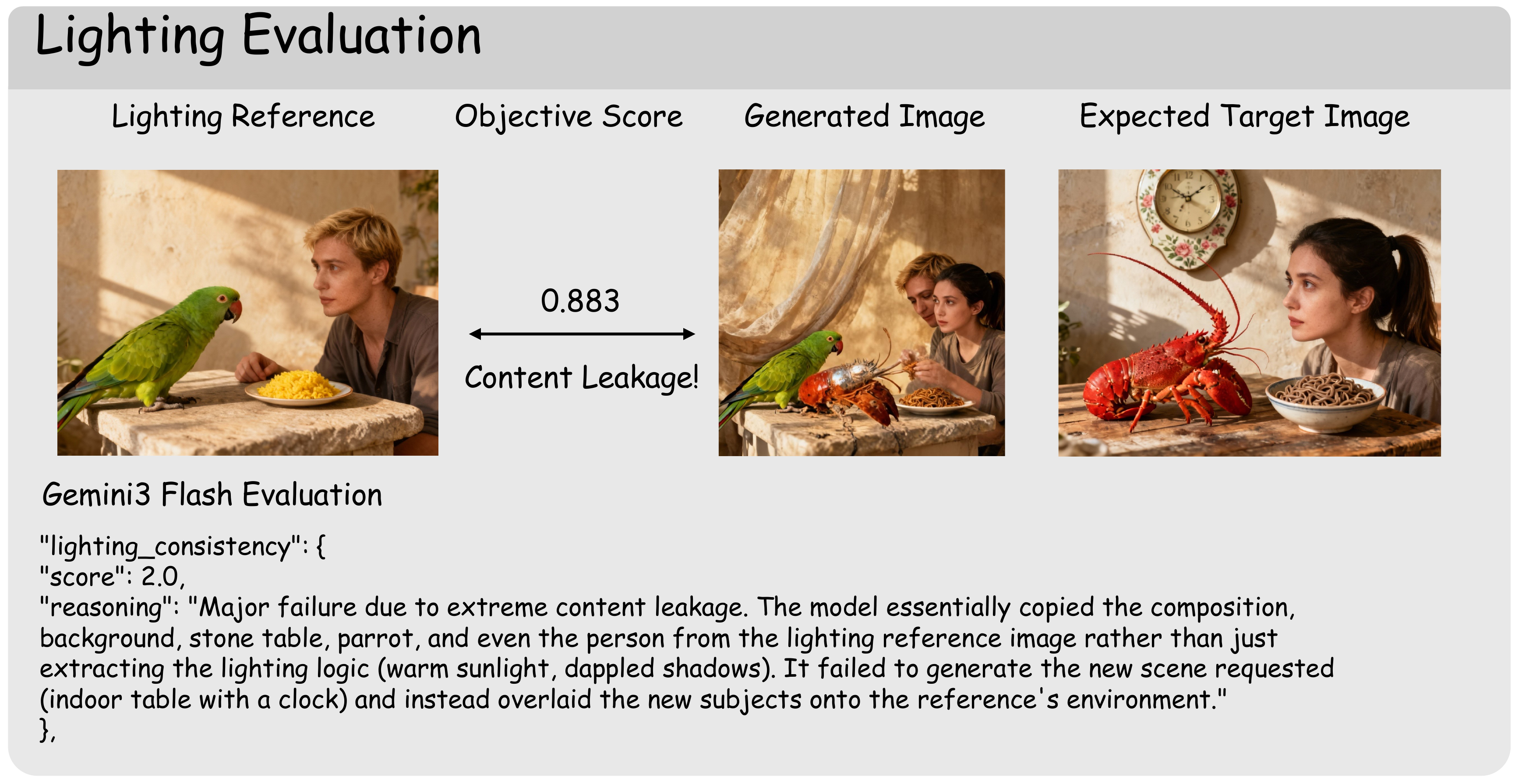}
\caption{Example of superficial alignment captured by objective metrics. The lighting objective metric erroneously assigns a high score due to content leakage from the reference image, while the MLLM evaluator identifies this flaw.}
\label{fig:eval_case_lighting}
\end{figure*}

\mypara{Prompt Reconstruction.}
The original T2I prompts are unsuitable for multi-reference generation task because they lack referential expressions to link specific concepts (e.g., subject, style) to their corresponding reference images. To bridge this gap, we employ Gemini 3 Flash to restructure the prompts. Specifically, we instruct the model to insert referential pointers (e.g., "in reference image X") immediately following the subjects. Furthermore, we replace explicit descriptions of background, lighting, pose, and style with concise referential pointers. This modification compels the model to derive these elements from the provided visual references rather than relying on textual prompts. The reference pointers for background, pose, lighting, and style are illustrated in Fig.~\ref{fig:reference pointer}.

\mypara{Data Filtering.}
Our data filtering pipeline integrates both automated checks and expert annotation.
First, for each reference type, we perform a separate automatic quality check, as detailed below.

\textbf{Subject reference.} We use Gemini 3 Flash to assess the consistency between each raw subject and modified subject pair.

\textbf{Background reference.} We use Gemini 3 Flash to check the image quality of the background reference and to check whether the removal of all specified foreground subjects is complete.

\textbf{Style reference.} We use Gemini 3 Flash to evaluate the style consistency between the stylized target image and its corresponding style reference.

\textbf{Lighting and pose reference.} We leverage their corresponding evaluation objective metrics detailed in Sec.~\ref{subsec:obj metrics} to assess their quality.

Subsequently, we invite a group of experts to go through the filtered images to better ensure that the quality of the images aligns with human preference.
\begin{figure*}[tbp]
\centering
\includegraphics[width=\linewidth]{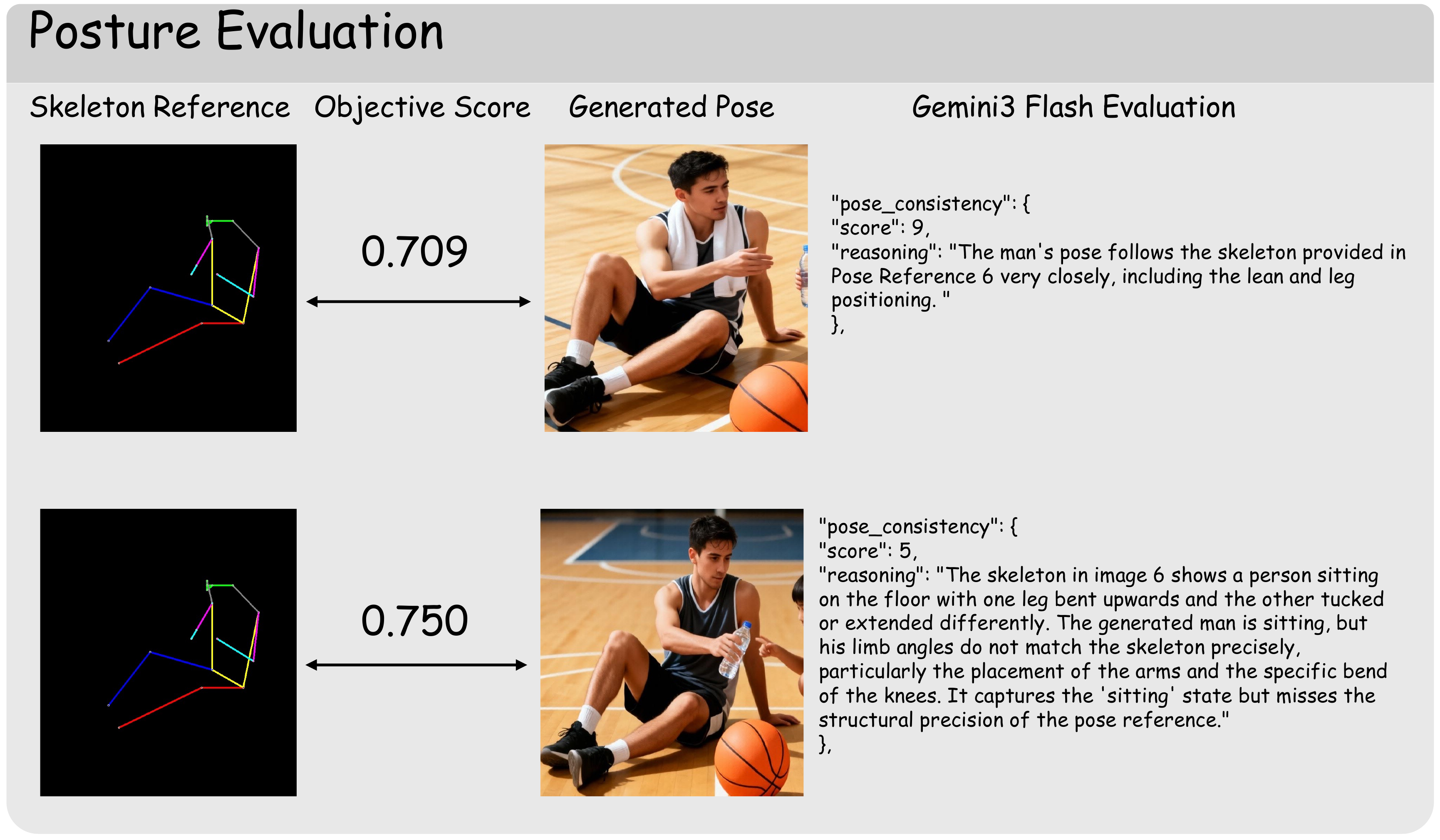}
\caption{Example case of the inconsistent scoring and lack of fine-grained spatial perception of MLLM-based evaluation. While the objective metric yields consistent scores for structurally similar poses, the MLLM evaluator exhibits significant variance (scoring 9 vs. 5) and provides inaccurate reasoning on the fine-grained pose alignment.}
\label{fig:eval_case_pose}
\end{figure*}

\subsection{Details of Objective Metrics}
\label{subsec:obj metrics}
We evaluate the generated images across five reference types using a combination of established and customized objective metrics. The details are as follows.

\mypara{Subject Consistency.} We first employ the Grounded-SAM2 model to segment all the subjects from the generated image based on the subject instance names used to generate that image. Then, following previous works~\cite{wang2025psr, wu2025less},  we calculate the pair-wise CLIP-I~\cite{radford2021learning} and DINOv2~\cite{oquab2023dinov2} scores between each segmented subject and its corresponding subject reference. We present the final subject consistency score by averaging the two scores.

\mypara{Background Consistency.} To focus on background fidelity, we mitigate the interference of foreground by masking subjects in both the generated image and its background reference (derived as described in Sec.~\ref{subsec:data construct}). Specifically, given a generated image, we first utilize the Grounded-SAM2 model to obtain the masks of the foreground subjects. Then, we apply the masks to both the generated image and its background reference. Inspired by VBench~\cite{huang2024vbench}, we compute the CLIP-I score between the two masked images to measure background consistency.

\mypara{Style Consistency.} We compute the established CSD~\cite{somepalli2024measuring} score between the generated image and its style reference to measure the style consistency. 

\mypara{Lighting Consistency.} As there are few works dedicated to develop a robust metric for measuring lighting consistency, we customize a metric considering the following dimensions: luminance, contrast, color temperature, lighting distribution, and shadow\&highlight. We implement the metric via the OpenCV library.

\mypara{Pose Consistency.} We first segment the target person using Grounded-SAM2 and extract pose skeletons via AlphaPose~\cite{alphapose}. Pose consistency is then quantified by averaging the Object Keypoint Similarity (OKS)~\cite{chen2020monocular}, Percentage of Correct Keypoints (PCK)~\cite{chen2020monocular}, and a customized angle similarity. The latter models skeleton segments as vectors and computes the mean cosine similarity across all corresponding vector pairs.

\subsection{Prompts for MLLM-based Evaluation}
\label{subsec:eval_prompt}
Existing MLLM-based evaluation for multi-reference image generation~\cite{oshima2025multibanana, xia2025dreamomni2} often overlooks the distinct characteristics of different reference types, resulting in undifferentiated and imprecise consistency evaluations. We tackle this by crafting specialized evaluation instructions for each reference type to guide the MLLM toward a more fine-grained analysis. Leveraging the capability of modern MLLMs (e.g., Gemini 3 series~\cite{team2023gemini}) to handle long-context visual inputs with high fidelity~\cite{oshima2025multibanana}, we implement a one-round inference process. This method feeds the generated image alongside the complete set of reference images and prompts into Gemini 3 Flash. This design not only ensures a comprehensive, holistic evaluation but also optimizes efficiency by minimizing API calls and processing time. We perform evaluation for the following 6 dimensions: subject consistency, background consistency, style consistency, lighting consistency, aesthetic, and prompt following. Compared with the objective metrics, we remove the pose consistency based on our observation detailed in Sec.~\ref{subsec:comparison of obj and mllm}. And we include aesthetic and prompt following to better leverage the high-level semantic understanding ability of the MLLM. The score for each dimension ranges from 0 to 10, and we present the overall results by averaging the 6 dimensions. The prompt for our MLLM-based evaluation is shown in \cref{fig:mllm system prompt,fig:mllm user prompt}.


\subsection{Limitations of Objective Metrics and MLLM Evaluation}
\label{subsec:comparison of obj and mllm}
As discussed in Sec.2.1 in the main paper, relying solely on objective metrics or MLLM-based evaluations presents inherent limitations. We provide two case studies on different reference types to illustrate the respective strengths and weaknesses of these two evaluation perspectives.

Objective metrics, which typically operate on fixed rules and low-level visual features, often struggle to comprehend high-level semantic constraints, making them susceptible to superficial alignment. This phenomenon is exemplified in Fig.~\ref{fig:eval_case_lighting}. In this instance, the generated image suffers from severe content leakage from the lighting reference. The model erroneously retains the man, the parrot, and the stone table from the reference image instead of only extracting the "lighting logic" (e.g., warm sunlight and dappled shadows). Consequently, the objective metric assigns a deceptively high score (0.883) due to the high pixel-level similarity. In contrast, the MLLM-based evaluator, guided by specific semantic instructions, successfully identifies this flaw, recognizing that the model failed to decouple the lighting effects from the reference scene's composition.

Conversely, MLLM-based evaluation often lacks the capacity for fine-grained spatial perception, leading to inconsistent scoring and unstable judgments when comparing visually similar results. This limitation is particularly evident in pose consistency evaluation, as illustrated in Fig.~\ref{fig:eval_case_pose}. When presented with two generated images featuring nearly identical poses, the MLLM’s scoring exhibits significant variance (9 vs. 5). In the first case (score 9), the MLLM fails to detect the clear structural discrepancies between the generated human figure and the reference skeleton. In the second case (score 5), however, it becomes overly critical of similar limb angle misalignments. In contrast, the objective metric provides a much more stable and granular assessment, yielding relatively consistent scores (0.709 vs. 0.750) that accurately reflect the similar level of pose alignment in both samples. This observation also leads to our removal of pose consistency assessment in MLLM-based evaluation.

In summary, combining objective metrics with MLLM-based evaluation bridges the gap between low-level visual fidelity and high-level semantic alignment. This hybrid approach compensates for individual limitations in spatial precision and semantic awareness, ensuring a more reliable evaluation framework.

%% file: sections/appdix_additional_exp.tex
\section{Additional Experiments and Analysis}
\label{sec:additional results}
\begin{table}[t]
\centering
\setlength{\tabcolsep}{6pt}  
\renewcommand{\arraystretch}{1.1}
\caption{Performance of various methods on DreamBench++ objective metrics. The best results are shown in \textbf{bold}.}
\begin{tabular}{lcccc}
\toprule
Methods                 & DINOv1 & DINOv2 & CLIP-I & CLIP-T \\ \midrule
OmniGen2~\cite{wu2025omnigen2}                & 0.56   & 0.54   & 0.77   & \textbf{0.35}   \\
BAGEL~\cite{deng2025bagel}                   & 0.44   & 0.43   & 0.72   & \textbf{0.35}   \\
FLUX.2 {[}klein{]} Base~\cite{flux-2-2025} & 0.49   & 0.48   & 0.74   & \textbf{0.35}   \\ \midrule
Qwen-Image-Edit-2511~\cite{wu2025qwenimagetechnicalreport}    & 0.52   & 0.51   & 0.75   & 0.34   \\\rowcolor{gray!15}
\hspace{0.5em}+Ours     & \textbf{0.60}   & \textbf{0.58}   & \textbf{0.78}   & \textbf{0.35}   \\ \bottomrule
\end{tabular}
\label{tab:dreambenchplus_obj}
\end{table}
\begin{figure}[htbp]
\centering
\includegraphics[width=0.8\linewidth]{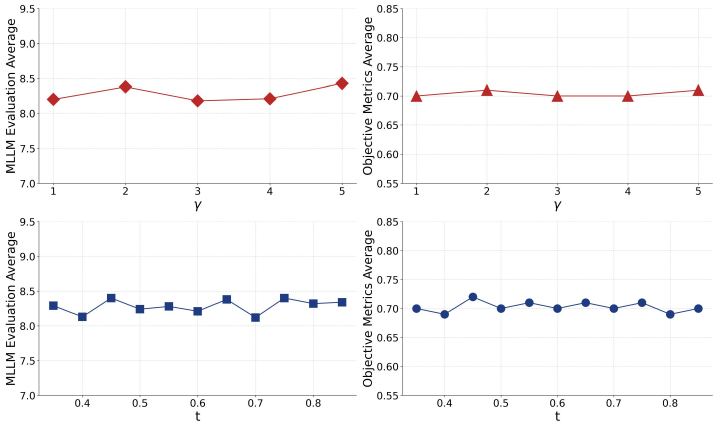}
\caption{Effect of $\gamma$ and $t$ on OmniRef-Bench. MLLM Evaluation Average and Objective Metrics Average denote the mean scores computed across all evaluation dimensions of MLLM evaluation and objective metrics in OmniRef-Bench.}
\label{fig:hyper_analysis}
\end{figure}
\subsection{More Results on DreamBench++}
\label{subsec:dreambench++_result}
To ensure our method enhances the base model's MRIG ability without compromising its basic image editing capability, we additionally evaluate it on the objective metrics of DreamBench++~\cite{peng2024dreambench}. As shown in Tab.~\ref{tab:dreambenchplus_obj}, our method outperforms mainstream open-source models on DreamBench++. Compared with the baseline model Qwen-Image-Edit-2511, it achieves improvements of 15.4\% and 4.0\% on the DINOv1 and CLIP-I metrics, respectively. These results indicate that our approach not only enhances model performance on complex MRIG tasks but also improves performance on single-image editing benchmarks, demonstrating the effectiveness and generalization capability of DyRef.

\subsection{Hyperparameter Sensitivity Analysis}
\label{subsec:hyper_analysis}
We conduct a sensitivity analysis on the main hyperparameters $t$ and $\gamma$ involved in our method, as shown in Fig.~\ref{fig:hyper_analysis}. When $\gamma$ is fixed at 2, the best performance is achieved at $t = 0.45$, followed by $t = 0.65$. When $t$ is fixed at 0.65, the optimal result is obtained at $\gamma = 5$, followed by $\gamma = 2$. For other values, model performance under objective metrics remains relatively stable, while the Gemini 3 Flash evaluation shows slight fluctuations. This may be attributed to variations in the initial noise and the additional noise introduced during training to ensure sample diversity. In addition, Gemini 3 Flash may also exhibit fluctuation in fine-grained scoring. Therefore, the results across both objective metrics and MLLM-based evaluations remain stable with respect to $t$ and $\gamma$, highlighting the robustness of our method.

\subsection{Motivation of Difficulty-aware Advantage Reweighting}
\label{subsec:reward distribution}
\begin{figure*}[t]
\centering
\includegraphics[width=\linewidth]{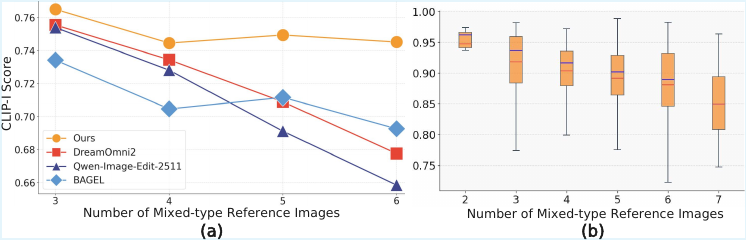}
\caption{\textbf{Motivation of Difficulty-aware Advantage Reweighting}. \textbf{(a)} The subject consistency of current open-source models on OmniRef-Bench decreases as the number of mixed-type reference images increases. \textbf{(b)} During training, the distribution range of CLIP-I rewards corresponding to different numbers of mixed-type reference images is narrow, and both the mean (red line) and the median (blue line) of the reward scores decrease as the number of mixed-type reference images increases.}
\label{fig:performance analysis}
\end{figure*}
As illustrated in Fig~\ref{fig:performance analysis}(a), evaluation on OmniRef-Bench shows that current mainstream open-source models perform poorly on complex MRIG tasks. Model performance deteriorates rapidly as the number of mixed-type reference images increases. To address this performance degradation, in the first stage, we use SFT to equip the model with the preliminary capability to handle complex MRIG tasks. In the second stage, we attempt to further align the model with human preferences using Flow-GRPO~\cite{liu2025flow}. 

However, as shown in Fig.~\ref{fig:performance analysis}(b), during Flow-GRPO training, the CLIP-I reward, which quantifies the overall semantic consistency between generated and target images, exhibits a narrow distribution concentrated between 0.75 and 0.95 (the theoretical range is -1 to 1). Moreover, this reward score declines as the number of mixed-type reference images increases. Motivated by these observations, we propose Difficulty-aware Advantage Reweighting (DAR) and Discriminative Reward Scaling (DRS). DAR computes the average reward for each group in Flow-GRPO and dynamically adjusts the contribution of each sample to the final loss based on this average reward. This encourages the model to enhance performance on samples with numerous mixed-type reference images that receive low rewards, while preventing overfitting on samples with fewer references and high rewards. DRS enlarges the reward differences across samples for better training.

\subsection{Detail Analysis of $F$}
\label{subsec:analysis_F}
\begin{table}[t]
\centering
\setlength{\tabcolsep}{2pt}  
\renewcommand{\arraystretch}{1.3}
\caption{Effect of different F on OmniRef-Bench. Bg, Aes, IF, and Avg are the abbreviations for background, aesthetic, instruction following, and average, respectively.}
\resizebox{\textwidth}{!}{%
\begin{tabular}{lccccccccccccc}
\toprule
\multirow{2.5}{*}{Methods} & \multicolumn{6}{c}{Objective Metrics}                                                         & \multicolumn{7}{c}{MLLM Evaluation}                                                                           \\ \cmidrule(lr){2-7} \cmidrule(lr){8-14} 
                         & Subject       & Style         & Bg            & Light         & Pose          & Avg           & Subject       & Bg            & Style         & Light         & Aes           & IF            & Avg           \\ \midrule
Qwen-Image-Edit-2511     & 0.55          & 0.15          & 0.84          & 0.71          & 0.71          & 0.59          & 6.50          & 4.35          & 1.61          & 6.81          & 6.09          & 4.46          & 4.97          \\
\hspace{0.5em}+Ours (F1)                & \textbf{0.63} & 0.44          & \textbf{0.88} & 0.74          & 0.83 & 0.70           & 8.36 & 8.57 & 8.52          & 8.44          & 8.10           & 8.08 & 8.34          \\
\hspace{0.5em}+Ours (F2)                & 0.62 & \textbf{0.46} & 0.87          & \textbf{0.75} & \textbf{0.84} & \textbf{0.71} & \textbf{8.41} & \textbf{8.61}          & \textbf{8.72} & \textbf{8.59} & \textbf{8.11} & \textbf{8.12} & \textbf{8.43} \\ \bottomrule
\end{tabular}%
}
\label{tab:analysis_F}
\end{table}
\begin{table}[t]
\centering
\setlength{\tabcolsep}{2pt}  
\renewcommand{\arraystretch}{1.3}
\caption{Effect of different visual semantic feature extraction models (CLIP and SigLIPv2) on OmniRef-Bench performance.}
\resizebox{\textwidth}{!}{%
\begin{tabular}{lccccccccccccc}
\toprule
\multirow{2.5}{*}{Methods} & \multicolumn{6}{c}{Objective Metrics}                                                         & \multicolumn{7}{c}{MLLM Evaluation}                                                                           \\ \cmidrule(lr){2-7} \cmidrule(lr){8-14} 
                         & Subject       & Style         & Bg            & Light         & Pose          & Avg           & Subject       & Bg            & Style         & Light         & Aes           & IF            & Avg           \\ \midrule
Qwen-Image-Edit-2511     & 0.55          & 0.15          & 0.84          & 0.71          & 0.71          & 0.59          & 6.50          & 4.35          & 1.61          & 6.81          & 6.09          & 4.46          & 4.97          \\
\hspace{0.5em}+Ours (CLIP)              & \textbf{0.63} & 0.42          & \textbf{0.89} & \textbf{0.75}          & 0.82 & 0.70           & \textbf{8.59} & \textbf{8.65} & 7.46          & 8.51          & 7.92          & 7.84 & 8.16          \\
\hspace{0.5em}+Ous (SigLIPv2)           & 0.62 & \textbf{0.46} & 0.87          & \textbf{0.75} & \textbf{0.84} & \textbf{0.71} & 8.41 & 8.61          & \textbf{8.72} & \textbf{8.59} & \textbf{8.11} & \textbf{8.12} & \textbf{8.43} \\ \bottomrule
\end{tabular}%
}
\label{tab:comparison_clip_siglipv2}
\end{table}
We explore two strategies to enhance the reward discriminability among samples within the same group of Flow-GRPO~\cite{liu2025flow}, denoted as $F_1$ and $F_2$. 
Let the training set be $\mathcal{S}$. For each sample $i \in \mathcal{S}$, let $p_i$ denote its reward score and $n_i$ represent the number of reference images associated with it. 
We compute the adjusted reward $\hat{p}_i$ as follows.
\begin{equation}
F_1: \quad \hat{p}_i = p_i^{\,t \cdot n_i},
\end{equation}
\begin{equation}
F_2: \quad \hat{p}_i = \text{Sigmoid}\big((k_1+k_2(n_i-n_0)) (p_i - t)\big),
\end{equation}
where $n_0$ denotes the minimum number of reference images in the training data, and $t (0<t<1)$ is a hyperparameter that controls the distribution of rewards. $k_1$ and $k_2$ are hyperparameters that control the influence of samples with different numbers of mixed-type reference images on the reward.
Both $F_1$ and $F_2$ enlarge the reward differences among samples within the same group. 
Moreover, groups associated with a larger number of mixed-type reference images receive stronger amplification of intra-group reward differences, thereby facilitating the learning of advantageous samples.
For both $F_1$ and $F_2$, we conduct experiments with a range of values for the hyperparameter $t$ and report the best-performing under $F_1$ and $ F_2$, respectively, as shown in Tab.~\ref{tab:analysis_F}. 
Overall, $F_2$ outperforms $F_1$. This may be because $F_2$ provides a more explicit reward shaping mechanism: rewards higher than the threshold $t$ are amplified, while those lower than $t$ are suppressed. 
Therefore, in all subsequent experiments, we adopt $F_2$ as the default configuration of Discriminative Reward Scaling (DRS).

\subsection{Comparison Between CLIP and SigLIPv2}
\label{subsec:comparision_clip_siglipv2}
As discussed in Sec. 3.2 of the main paper, we introduce a reward that measures the overall semantic consistency between generated and target images. This reward is computed by extracting semantic features from the generated and target images using a visual encoder and then calculating their cosine similarity as the reward signal for training. Previous works~\cite{peng2024dreambench,chen2025multiref} usually use CLIP to extract semantic features from images. In addition, we further investigate the effectiveness of SigLIPv2. As shown in Tab.~\ref{tab:comparison_clip_siglipv2}, we observe that semantic features extracted by SigLIPv2 lead to better training results. This may be because SigLIPv2 performs better as the visual encoder in vision language models and provides more comprehensive and informative semantic representations. Therefore, in subsequent experiments, we use SigLIPv2 to extract semantic features to compute the consistency reward between generated and target images.

\subsection{Additional Results on OmniContext and MultiBanana}
\label{subsec:omnicontext_multibanana_results}
\begin{table}[t]
\centering
\setlength{\tabcolsep}{2pt}  
\renewcommand{\arraystretch}{1.1}
\caption{Performance of various methods on OmniContext. "Char.+Obj." indicates Character+Object. The best and second best results are shown in \textbf{bold} and \underline{underline}, respectively. Qwen-2511 denotes the Qwen-Image-Edit-2511.}
\resizebox{\textwidth}{!}{%
\begin{tabular}{lccccccccc}
\toprule
\multirow{2.5}{*}{Methods} & \multicolumn{2}{c}{SINGLE}    & \multicolumn{3}{c}{MULTIPLE}                  & \multicolumn{3}{c}{SCENE}                     & \multirow{2.5}{*}{Average} \\ \cmidrule(lr){2-3} \cmidrule(lr){4-6} \cmidrule(lr){7-9}
                        & Character     & Object        & Character     & Object        & Char.+Obj.    & Character     & Object        & Char.+Obj.     &                          \\ \midrule
BAGEL~\cite{deng2025bagel}                   & 6.99          & 5.94          & 5.20          & 6.74          & 7.02          & 4.75          & 4.59          & 5.55          & 5.85                     \\
OmniGen2~\cite{wu2025omnigen2}                & 8.16          & 8.18          & 7.43          & 7.29          & 7.88          & 7.34    & 6.62          & 7.37          & 7.54                     \\
DreamOmni2~\cite{xia2025dreamomni2}              & 7.56          & 7.49          & 6.53          & 6.53          & 6.69          & 6.05          & 5.86          & 6.09          & 6.60                      \\ \midrule
Qwen-2511~\cite{wu2025qwenimagetechnicalreport}               & 8.56    & 8.44    & \textbf{8.38} & \textbf{8.88} & 7.91    & 6.84          & \textbf{8.04} & \textbf{8.10} & 8.14               \\ \rowcolor{gray!15}
\hspace{0.5em}+Ours                    & \textbf{8.59} & \textbf{8.80} & 8.24    & 8.80    & \textbf{8.09} & \textbf{7.59} & 7.93    & 7.89    & \textbf{8.24}            \\ \bottomrule
\end{tabular}%
}
\label{tab:omnicontext}
\end{table}
To further verify that our method remains effective beyond the complex MRIG scenarios introduced in OmniRef-Bench, we evaluate it on two simpler MRIG benchmarks, OmniContext~\cite{wu2025omnigen2} and MultiBanana~\cite{oshima2025multibanana}. As shown in Tab.~\ref{tab:omnicontext} and Tab.~\ref{tab:multibanana}, compared with the baseline Qwen-Image-Edit-2511 (Qwen-2511), our method improves the average performance across all metrics by 1.2\% and 14.8\% on OmniContext and MultiBanana, respectively, demonstrating its effectiveness and strong generalization ability.

\begin{table}[t]
\centering
\setlength{\tabcolsep}{6pt}  
\renewcommand{\arraystretch}{1.1}
\caption{Performance of various methods on MultiBanana (10-point scale; the higher the better). Following the official evaluation protocol provided by the benchmark authors, we use Gemini-2.5 and GPT-5 as evaluators.}
\resizebox{\textwidth}{!}{%
\begin{tabular}{lccccccc}
\toprule
Methods    & Single        & Two           & X Objects     & X-1 + Local   & X-1 + Global  & X-1 + Background & Average       \\ \midrule
DreamOmni2 & \textbf{6.52} & 4.07          & 2.80           & 3.04          & 2.87          & 2.59             & 3.65          \\
OmniGen2   & 5.92          & 3.44          & 3.26    & 3.60     & 3.37   & 3.02       & 3.77          \\ \midrule
Qwen-2511  & 6.37    & \textbf{4.54} & 3.01          & 2.93          & 3.18          & 2.65             & 3.78    \\
\hspace{0.5em}+Ours     & 6.34          & 4.42    & \textbf{3.96} & \textbf{3.65} & \textbf{4.04} & \textbf{3.63}    & \textbf{4.34} \\ \bottomrule
\end{tabular}%
}
\label{tab:multibanana}
\end{table}

\subsection{Generalization to More Reference Types and User Study}
\label{subsec:more_ref_user}
To further demonstrate the generalization capability of our method, we expand the user study to include 40 human experts and 50 randomly selected evaluation samples. In addition, we evaluate our method on four reference types that are not included in the training data. As shown in Tab.~\ref{tab:more_user_study}, our method consistently outperforms the baseline methods under this larger-scale human evaluation setting.
We further assess inter-rater agreement among evaluators using Kendall’s W, which quantifies the degree of consistency in rankings and judgments across annotators. The high agreement score (mean Kendall’s W = 0.78) indicates strong consistency among human evaluators.

As shown in Tab.~\ref{tab:more_ref_type}, we further evaluate our method on MultiBanana~\cite{oshima2025multibanana} and MultiRef~\cite{chen2025multiref}, which include additional reference types beyond those encountered during training. The results demonstrate that our method consistently improves performance on these unseen reference types, highlighting its strong generalization ability beyond the training distribution.

\begin{table}[t]
\centering
\setlength{\tabcolsep}{6pt}  
\renewcommand{\arraystretch}{1.1}
\caption{More user study.  we expand the user study to 50 samples with 40 human expert evaluators. PLCC, SROCC, and KROCC denote Pearson, Spearman, and
Kendall correlation coefficients, respectively. Kendall’s W quantifies the degree of consistency among different human annotators in their rankings or judgments.}
\resizebox{\textwidth}{!}{%
\begin{tabular}{lcccccccl}
\toprule
\multirow{2.5}{*}{Evaluation Setup} & \multicolumn{4}{c}{Evaluation Results}                   & \multicolumn{3}{c}{Correlation Analysis}                              & Inter-rater agreement                     \\ \cmidrule(lr){2-5} \cmidrule(lr){6-8} \cmidrule(lr){9-9}
                                  & Ours          & Nano Pro      & Seedream 4.5 & Qwen-2511 & PLCC                  & SROCC                 & KROCC                 & Mean Kendall's W                          \\ \midrule
Gemini3 Flash                     & 8.38    & \textbf{8.50} & 8.13         & 4.97      & \multirow{2}{*}{0.89} & \multirow{2}{*}{0.80} & \multirow{2}{*}{0.67} & \multicolumn{1}{c}{\multirow{2}{*}{0.78}} \\
Human                             & \textbf{3.53} & 2.96    & 2.30         & 1.21      &                       &                       &                       & \multicolumn{1}{c}{}                      \\ \bottomrule
\end{tabular}%
}
\label{tab:more_user_study}
\end{table}

\begin{table}[t]
\centering
\setlength{\tabcolsep}{6pt}  
\renewcommand{\arraystretch}{1.1}
\caption{More reference types. We further evaluate DyRef on four reference types that are absent from the training set. Compared with the baseline, DyRef consistently outperforms it across all four unseen reference types, demonstrating strong generalization to diverse reference conditions beyond those encountered during training.}
\begin{tabular}{lcccc}
\toprule
\multirow{2.5}{*}{Methods} & \multicolumn{2}{c}{MultiBanana} & \multicolumn{2}{c}{MultiRef}  \\ \cmidrule(lr){2-3} \cmidrule(lr){4-5} 
                         & Text           & Tone           & bbox          & semantic map  \\ \midrule
Qwen-2511                & 3.52           & 3.53           & 7.12          & 8.79          \\
\hspace{0.5em}+Ours                   & \textbf{3.82}  & \textbf{3.83}  & \textbf{8.71} & \textbf{8.86} \\ \bottomrule
\end{tabular}
\label{tab:more_ref_type}
\end{table}

%% file: sections/appdix_train_details.tex
\section{Training Details}
\label{sec:training details}
\mypara{In the First Stage.}
Based on existing multi-reference image generation (MRIG) models, such as Qwen-Image-Edit-2511~\cite{wu2025qwenimagetechnicalreport} and FLUX.2-klein-base~\cite{flux-2-2025}, we first perform supervised fine-tuning using LoRA~\cite{hu2022lora} with the Flow-Matching loss~\cite{lipman2022flow}.
Specifically, we fine-tune only the MMDiT module responsible for image generation, while keeping all other components frozen. Based on the data construction pipeline we designed in Sec.~\ref{subsec:data construct}, we curate approximately 14000 high-quality training samples. During training, the LoRA rank is set to 64 and the learning rate is set to $1\times10^{-4}$. The model is trained for 5 epochs using the AdamW~\cite{loshchilov2017decoupled} optimizer with a weight decay of 0.01.

\mypara{In the Second Stage.}
To further narrow the gap between open-source models and closed-source models, we perform a second stage of training initialized from the SFT checkpoint. Specifically, we employ Flow-GRPO~\cite{liu2025flow} to further align the model with human preferences, and we introduce Difficulty-aware Advantage Reweighting (DAR) and Discriminative Reward Scaling (DRS) to address existing issues in Flow-GRPO training (details in Sec.~\ref{subsec:reward distribution}). Similar to the first stage, we keep all other components frozen and optimize only the MMDiT module.

Moreover, we employ a group size $G=24$, a noise level $a=1.0$, and an image resolution of $512 \times 512$. The KL ratio $\beta$ is fixed at $0.04$. For parameter-efficient fine-tuning, we adopt LoRA with rank $r=64$ and scaling factor $\alpha=64$. Training is conducted for 200 optimization steps, using a global batch size of 1 and a learning rate of $3 \times 10^{-4}$. We utilize the Adam~\cite{diederik2014adam} optimizer $(\beta_1=0.9, \beta_2=0.999, \epsilon=1\times10^{-8})$ with a weight decay of $1\times10^{-4}$.
Inspired by GDPO~\cite{liu2026gdpo}, we normalize the advantages within each group by subtracting the mean and dividing by the standard deviation to better facilitate multi-reward learning. We further adopt SigLIPv2 and CSD as reward signals to improve the alignment between the generated data distribution and the target data distribution, while also enhancing stylistic fidelity. For DRS, we set $t=0.65$, $k_1=10$, and $k_2=3$, respectively. For DAR, the $\gamma$ is set to 5.

%% file: sections/appdix_eval_details.tex
\section{Evaluation Details}
\label{sec:eval details}
\subsection{Benchmarks and Baselines}
\label{subsec:bench details}
To demonstrate the effectiveness of DyRef, in addition to OmniRef-Bench for evaluating multi-reference image generation (MRIG) tasks, we further evaluate our method on two additional single-image editing benchmarks. In this section, we provide additional details on the benchmarks referenced in the main paper.

\mypara{DreamBench++.}
DreamBench++~\cite{peng2024dreambench} provides an automated evaluation framework that aligns well with human judgments. The benchmark contains samples with diverse types and varying difficulty levels and significantly surpasses the original DreamBench in both dataset scale and diversity, offering a more comprehensive evaluation standard.

\mypara{ImgEdit.}
ImgEdit~\cite{ye2025imgedit} contains 1.2 million carefully curated image editing pairs, covering diverse and complex single-turn editing tasks as well as challenging multi-turn scenarios. The dataset is organized into two major categories: single-turn editing and multi-turn editing. Single-turn editing includes local editing operations such as addition, removal, replacement, modification, motion changes, and object extraction, global editing operations such as background replacement and style or tone transformation, reference-based visual editing, and hybrid editing that applies two local editing operations to multiple objects simultaneously. Multi-turn editing involves three types of interactions: content memory, content understanding, and version backtracking, which are designed to support complex editing requirements in real-world applications.

\subsection{Backbones and Baselines}
\label{subsec:baselies}
In Sec. 4.1 of the main paper, we further validate the effectiveness and generalization ability of our method on two widely used open-source MRIG models: Qwen-Image-Edit-2511~\cite{wu2025qwenimagetechnicalreport} and FLUX.2-klein-base~\cite{flux-2-2025}.
In addition, we compare our approach with several mainstream methods for MRIG tasks. The details of these methods are as follows:

\mypara{Qwen-Image-Edit-2511.}
Qwen-Image-Edit-2511~\cite{wu2025qwenimagetechnicalreport} is an enhanced version of Qwen-Image-Edit-2509, improving capabilities in multi-person portrait consistency, geometric reasoning, and character consistency. Qwen-Image is a strong text-to-image generation model, particularly excelling at Chinese text rendering. It adopts a standard double-stream MMDiT architecture, where the input representations are provided by a frozen Qwen2.5-VL~\cite{Qwen2.5-VL} encoder and a VAE~\cite{kingma2013auto} encoder. Qwen-Image-Edit is a single-image editing model obtained through post-training based on Qwen-Image. Building upon this model, Qwen-Image-Edit-2509 further extends the capability of Qwen-Image-Edit to support multi-reference image generation.


\mypara{FLUX.2 [klein].}
FLUX.2 [klein]~\cite{flux-2-2025} integrates image generation and editing within a compact architecture, delivering state-of-the-art image quality with end-to-end inference latency of less than one second. It is designed for applications that require real-time image generation without sacrificing quality. The model supports multi-reference image generation and can run on consumer-grade hardware equipped with only 13 GB of GPU memory.

\mypara{OmniGen2.}
OmniGen2~\cite{wu2025omnigen2} is a unified open-source multimodal generative model that supports text-to-image generation, image editing, and in-context generation in a single framework. Compared with OmniGen~\cite{xiao2025omnigen}, it adopts decoupled decoding pathways for text and image modalities, using a separate autoregressive transformer for text and a diffusion transformer for image generation. It further employs two distinct image encoders, where ViT features are used for multimodal understanding, and VAE features are injected into the diffusion branch for image synthesis, enabling the model to better preserve text understanding capability while improving visual generation quality. In addition, OmniGen2 is trained with dedicated data construction pipelines for image editing and in-context generation, and also explores a reflection mechanism for iterative refinement. Despite its relatively modest parameter scale, OmniGen2 demonstrates competitive performance across multiple benchmarks, especially showing strong consistency among open-source models on in-context generation tasks.

\mypara{BAGEL.}
BAGEL~\cite{deng2025bagel} is a unified, decoder-only open-source multimodal foundation model that natively supports both the understanding and generation of text, images, and videos. To efficiently process diverse multimodal information, BAGEL employs a Mixture-of-Transformer-Experts (MoT) architecture featuring two distinct expert pathways: one dedicated to multimodal understanding and another specifically for multimodal generation. These experts operate on the same token sequence through a shared self-attention mechanism at every layer, mapping all major modalities into a unified token space. Pre-trained on trillions of tokens sourced from large-scale interleaved text, image, video, and web data, BAGEL exhibits emergent capabilities in complex multimodal reasoning, free-form visual manipulation, and multi-image compositional generation. Evaluation results on various benchmarks show that it serves as a powerful unified baseline, achieving text-to-image synthesis and editing quality competitive with top-tier specialist generators such as SD3~\cite{esser2024scaling} and FLUX.1-dev~\cite{flux2024}.

\mypara{DreamOmni2.}
DreamOmni2~\cite{xia2025dreamomni2} is a unified framework designed for multimodal instruction-based image editing and generation that extends customizability beyond concrete objects to support abstract attributes. It addresses the ambiguity of language-only instructions by natively processing multiple reference images alongside text prompts. To achieve this, the framework implements a novel index encoding and position encoding shift scheme; the index encoding helps the model identify specific image references (e.g., "Image 1" vs. "Image 2"), while the position offset prevents pixel confusion and copy-and-paste artifacts. Additionally, DreamOmni2 employs a joint training strategy that integrates a Vision-Language Model (VLM) to interpret complex, irregular user instructions and translate them into a structured format for the core generation model. 

\mypara{Nano Banana Pro.}
Nano Banana Pro~\cite{Nanobanana} is a closed-source image generation and editing model built on Gemini 3 Pro. It supports natural-language-driven image synthesis and image editing, and is designed to produce more accurate, context-rich visuals by leveraging Gemini’s reasoning and real-world knowledge. It also emphasizes strong text rendering, multilingual legible text generation, high-fidelity outputs, and more precise control for complex creative tasks such as infographics, diagrams, and other structured visual content, making it a strong closed-source baseline for instruction-guided image generation and editing. 

\mypara{Seedream 4.5.}
Seedream 4.5~\cite{seedream2025seedream} is a unified multimodal foundation model that integrates text-to-image generation and instruction-based image editing. It is particularly distinguished by its robust prompt adherence, layout-aware spatial reasoning, and exceptional capability in rendering precise typography and dense text. Owing to these strengths, Seedream 4.5 serves as a strong closed-source baseline in our experiments.

\subsection{Details of User Study}
\label{subsec:user study}

\mypara{Dataset Construction.} 
To verify the reliability of the MLLM-based scoring mechanism under complex multi-reference conditions, we utilize stratified sampling to select test cases. These samples uniformly cover 10 distinct combinations of reference image types. Each combination contains 2 to 3 test cases, ensuring the diversity and balance of the evaluation scenarios.

\mypara{Evaluation Metrics Details.} 
The five dimensions evaluated by the human experts are defined as follows: (1) \textit{Text Fidelity} requires the image to accurately reflect the prompt without semantic omissions. (2) \textit{Reference Similarity} demands high consistency with the reference images in identity and core details. (3) \textit{Composition Quality} assesses the spatial layout and natural blending among the subject, background, and other elements. (4) \textit{Visual Appeal} evaluates overall aesthetic quality, including clarity and the absence of generative artifacts. (5) \textit{Task Completion} strictly requires the simultaneous satisfaction of all constraints; failure in any single control dimension (e.g., correct pose but degraded background) results in a substantial penalty.

\mypara{Scoring Mechanism.} 
To quantify the discrepancy between human intuition and MLLM scoring, we employ two distinct mechanisms. For human evaluation, expert rankings are converted into relative scores, where the first place receives 4 points and the fourth place receives 1 point. The final human score is the average score across all experts. In contrast, the MLLM directly assigns an absolute score ranging from 1 to 10 based on predefined evaluation criteria.

\subsection{Reproducibility}
\label{subsec:reproducibility}
\mypara{Implementation Details.}
For OmniGen2, BAGEL, DreamOmni2, Qwen-Image-Edit-2511, and FLUX.2 [klein], we keep all hyperparameters consistent with those reported in the original papers or provided in their official code repositories. For DyRef, we use LoRA to fine tune the MMDiT component in both stages, with the rank set to 64. In the second stage, the parameter $t$ is set to 0.65 for DRS and $\gamma$ is set to 5 for DAR. During evaluation, the initial noise level is set to 1 and the number of denoising steps is set to 20, while other methods are evaluated with their default configurations.

\mypara{Experimental Code.}
To ensure transparency and reproducibility, all code, datasets, and detailed tutorials will be made publicly available soon.

%% file: sections/appdix_case_study.tex
\section{Case Study}
\label{sec:case study}

\begin{figure}[htbp]
\centering
\includegraphics[width=0.8\linewidth]{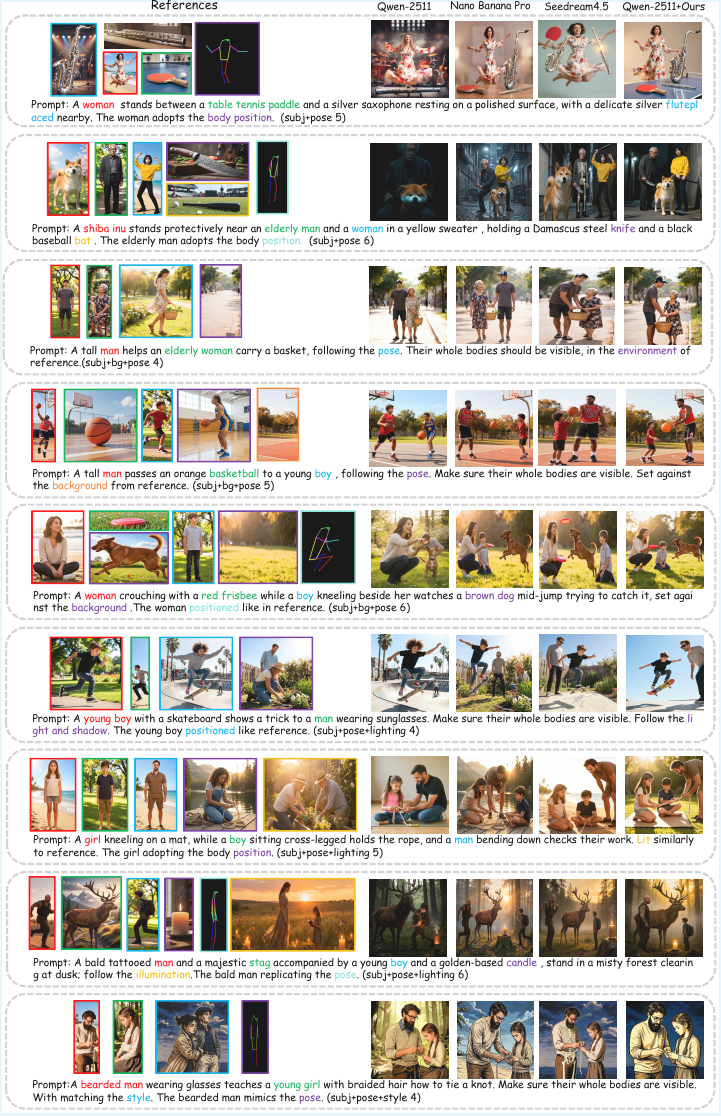}
\caption{Multi-reference image generation results of different methods. We compare our method with a range of state-of-the-art open-source and close-source models across diverse representative cases.}
\label{fig:case3_1}
\end{figure}

\begin{figure}[htbp]
\centering
\includegraphics[width=0.8\linewidth]{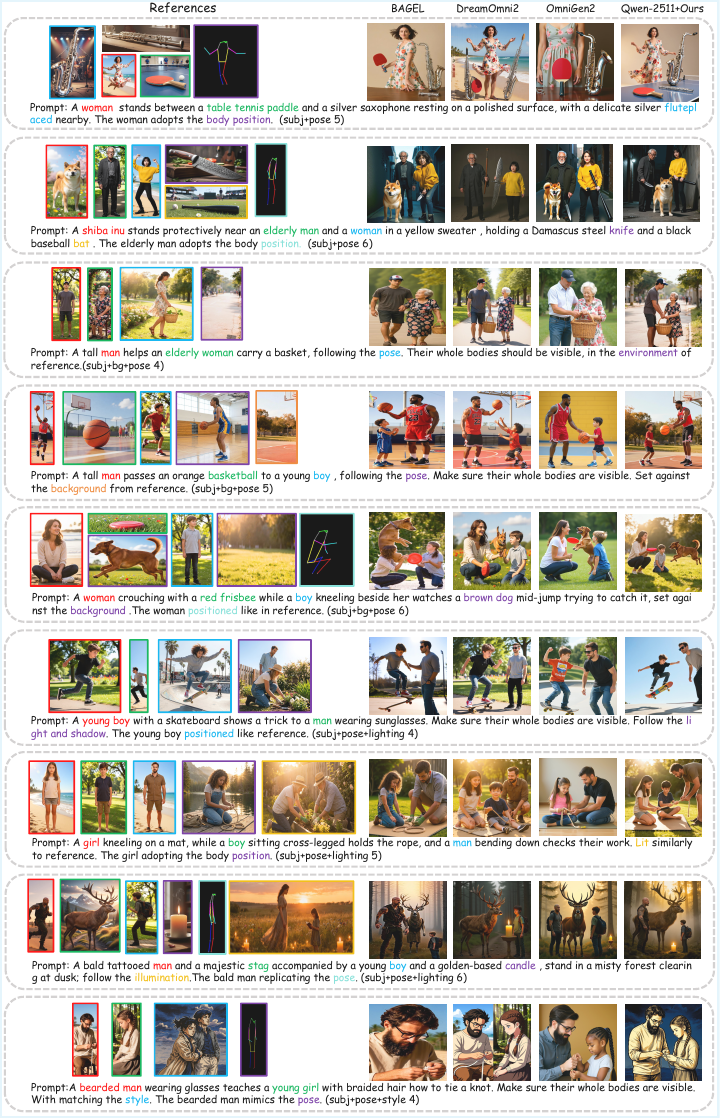}
\caption{Multi-reference image generation results of different methods. We compare our method with a range of state-of-the-art open-source and close-source models across diverse representative cases.}
\label{fig:case3_2}
\end{figure}

\begin{figure}[htbp]
\centering
\includegraphics[width=0.8\linewidth]{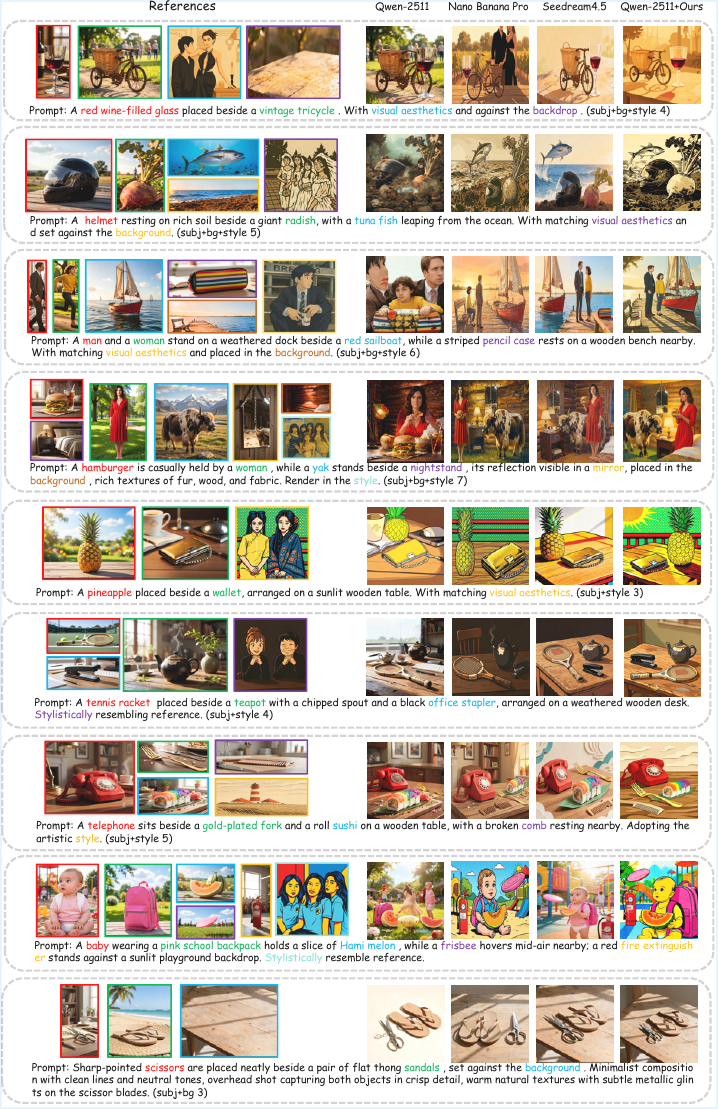}
\caption{Multi-reference image generation results of different methods. We compare our method with a range of state-of-the-art open-source and close-source models across diverse representative cases.}
\label{fig:case1_1}
\end{figure}

In this section, we conduct a qualitative analysis of different methods on the multi-reference image generation (MRIG) task. As shown in \cref{fig:case3_1,fig:case3_2,fig:case1_1}, in complex MRIG scenarios, the proposed method not only significantly outperforms mainstream open-source models such as Qwen-Image-Edit-2511, BAGEL, and DreamOmni2, but also achieves performance comparable to or even better than closed-source models including Nano Banana Pro and Seedream 4.5. These results demonstrate the effectiveness and robustness of the DyRef framework.

We further analyze the limitations of our method. As illustrated in Fig~\ref{fig:failed_case}, although the generated images exhibit good clarity and overall visual quality, limitations remain in instruction following. Specifically, the model either omits certain reference information or overuses it. For example, in Fig~\ref{fig:failed_case}(a), the character from the style reference image incorrectly appears in the generated image, indicating an over-referencing issue. In contrast, Fig~\ref{fig:failed_case}(b) and Fig~\ref{fig:failed_case}(c) fail to accurately reproduce the background and pose specified in the instruction, indicating reference omission.
These limitations may stem from the lack of explicit feature disentanglement across different reference types and the absence of a one-to-one alignment mechanism between instructions and reference information. As a result, the model may confuse different reference types and incorrectly aggregate features from unrelated images. These observations suggest that complex MRIG remains an open problem that requires further investigation.


\clearpage 
\begin{center}
\includegraphics[width=0.8\textwidth]{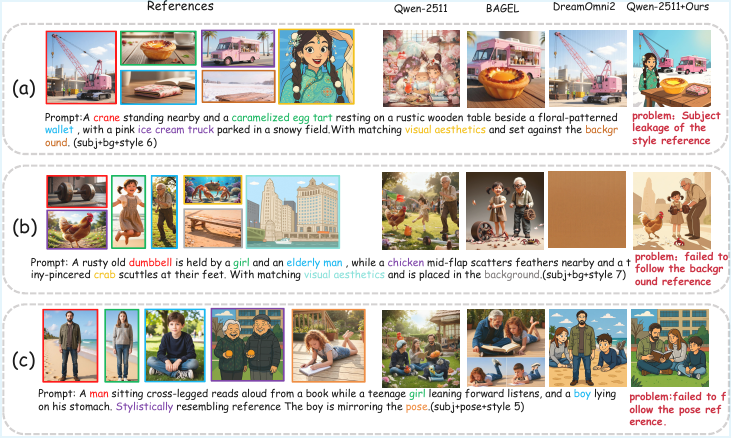}
\captionof{figure}{Failure cases of DyRef on the multi-reference image generation task.}
\label{fig:failed_case}
\end{center}

\vspace{6pt}  

\lstset{
  basicstyle=\small\ttfamily,
  backgroundcolor=\color{white},
  frame=none,
  breaklines=true,
  columns=flexible,
  keepspaces=true,
  xleftmargin=2em
}


\begin{figure}[tbp]
\begin{tcolorbox}[
  enhanced,
  breakable,
  colback=white,
  colframe=black!60,
  boxrule=0.6pt,
  arc=1pt,
  outer arc=1pt,
  toptitle=3pt,
  bottomtitle=3pt,
  coltitle=white,
  colbacktitle=headerGray,
  fonttitle=\bfseries\sffamily,
  title={System Prompt: Subject Instance Generation},
  left=8pt, right=8pt, top=6pt, bottom=6pt,
  fontupper=\small
]

\textbf{\large [System Prompt]}

\medskip
You are a realistic prompt generator for text-to-image. Output only the prompts.

\medskip
\textbf{Role:} Please be very realistic and generate 20 brief subject prompts for text-to-image generation.

\medskip
\textbf{Follow these rules:}
\begin{enumerate}[leftmargin=1.5em, itemsep=2pt]
  \item You will be given an \texttt{<asset category>}, you need to create an asset (brief subject prompt) based on the \texttt{<asset category>}.
  \item These descriptions can refer only to appearance descriptions/or to certain brands. e.g., ``Elon Musk in pajamas'', ``a tiger in a black hat'', ``A Mercedes sports car'', ``A blonde'', ``A door red on the left and green on the right''.
  \item Focus on the given \texttt{<asset category>} ONLY. Avoid adding separate accessories or objects.
  \item Do not repeat each asset, you need to use your logic and common sense of life to create.
  \item No more than 12 words for each asset.
\end{enumerate}

\medskip
\textbf{Example}

\medskip
\texttt{[asset category]:} Book

\medskip
\textbf{Output:}
\begin{itemize}[leftmargin=1.5em, itemsep=1pt, label={}]
  \item \texttt{[asset1]:} A book with a green cover
  \item \texttt{[asset2]:} comic book
  \item \texttt{[asset3]:} math book
  \item \texttt{[asset4]:} An open book
  \item \texttt{[asset5]:} Rotten books
  \item \texttt{[asset6]:} The book with ``love and power'' on the cover
  \item \texttt{[asset7]:} A book with a blue key on it
  \item \quad $\vdots$
  \item (Up to \texttt{[asset20]})
\end{itemize}

\medskip
\textbf{User:}

\medskip
\texttt{[asset category]:} \{category\}

\end{tcolorbox}
\caption{System prompt for subject instance generation. For every subject category in Objects365~\cite{shao2019objects365}, we leverage DeepSeek-v3~\cite{liu2024deepseek} to generate 20 concrete subject instances.}
\label{fig:subject instance generation prompt}
\end{figure}

\newpage
\begin{tcolorbox}[
  enhanced,
  breakable,
  colback=white,
  colframe=black!60,
  boxrule=0.6pt,
  arc=1pt,
  outer arc=1pt,
  toptitle=3pt,
  bottomtitle=3pt,
  coltitle=white,
  colbacktitle=headerGray,
  fonttitle=\bfseries\sffamily,
  title={System Prompt: T2I Prompt Generation},
  left=8pt, right=8pt, top=6pt, bottom=6pt,
  fontupper=\small
]

\textbf{\large [System Prompt]}

\medskip
\textbf{Role:}
You are a prompt composer for text-to-image generation. Combine multiple subject phrases into one coherent, vivid, and conflict-free prompt suitable for modern diffusion models (e.g., SDXL, Midjourney, Flux).

\medskip
\textbf{Tasks:}
\begin{enumerate}[leftmargin=1.5em, itemsep=2pt]
  \item Merge and prioritize attributes across all subjects; resolve conflicts logically.
  \item Clarify relationships, composition, and focal hierarchy (primary subject vs.\ secondary).
  \item Add tasteful art direction: lighting, environment, background, camera, lens, shot type, color palette, mood, time of day, material details, and post-processing.
  \item Preserve each subject's core identity; distribute attributes sensibly (don't duplicate or contradict).
  \item For each person subject, specify a \emph{diversified} pose ranging from static to dynamic (e.g., sitting with legs crossed, jumping with arms outstretched).
  \item ...
\end{enumerate}

\medskip
\textbf{Input format:}
\begin{itemize}[leftmargin=1.5em, itemsep=1pt]
  \item \texttt{subject\_phrases}: a list of short noun phrases describing subjects or objects.
  \item \texttt{variants}: number of alternative prompts to produce (default: 1).
\end{itemize}

\medskip
\textbf{Output format:}
Produce exactly $N$ variants ($N$ = \texttt{variants}). For each variant, output a single cohesive prompt ready for a text-to-image model.

\medskip
\textbf{Output requirements:} Return ONLY the prompt text itself.

\medskip
\textbf{Example}

\medskip
\textbf{Input:}\\
\texttt{subject\_phrases}: [\texttt{"A woman with neon pink hair"}, \texttt{"Rustic farmhouse desk with chipped paint"}]\\
\texttt{variants}: 1

\medskip
\textbf{Output:}\\
\textit{A neon-pink-haired woman seated with legs crossed at a rustic farmhouse desk with chipped paint, sunlit studio corner, soft window light kissing worn wood grain, gentle filmic tones; medium portrait at eye level, shallow depth of field, warm neutrals with a pop of neon accent in hair and stationery.}

\end{tcolorbox}
\captionof{figure}{System prompt for T2I prompt generation. We leverage Gemini 3 Flash~\cite{team2023gemini} to integrate logically-consistent combinations of subject instances into a coherent scene description, which is then formulated as the final T2I prompt.}
\label{fig:prompt for t2i prompt generation}

\begin{tcolorbox}[
  enhanced,
  breakable,
  colback=white,
  colframe=black!60,
  boxrule=0.6pt,
  arc=1pt,
  outer arc=1pt,
  toptitle=3pt,
  bottomtitle=3pt,
  coltitle=white,
  colbacktitle=headerGray,
  fonttitle=\bfseries\sffamily,
  title={System Prompt: MLLM-based Evaluation},
  left=8pt,
  right=8pt,
  top=6pt,
  bottom=6pt,
  fontupper=\small
]
\newpage
{\large\textbf{[System Prompt]}}

\medskip
\noindent You are an expert AI Art Director and Visual Quality Assurance Specialist. Your task is to evaluate a ``Generated Image'' against a ``Text Prompt'' and a set of ``Reference Images'' (Control Signals).

\medskip
\noindent You must assess the consistency of the generated image across 6 dimensions:

\begin{enumerate}[leftmargin=2em, itemsep=2pt, topsep=4pt]
  \item \textbf{Prompt Following}: How well does the image match the text description?
  \item \textbf{Subject Consistency}: Identity preservation (ignoring style).
  \item \textbf{Style Consistency}: Art style, medium, and texture match.
  \item \textbf{Lighting Consistency}: Illumination direction, color, and mood.
  \item \textbf{Background Consistency}: Environment and setting match.
  \item \textbf{Aesthetic}: Overall visual appeal and technical quality.
\end{enumerate}

\bigskip
\noindent\textbf{General Rules:}

\begin{itemize}[leftmargin=1.5em, itemsep=2pt, topsep=4pt]
  \item \textbf{Conservative Scoring}: Adopt a strict, critical mindset. Do not inflate scores.

  \begin{description}[
      labelwidth=6.5em,
      leftmargin=7em,
      labelsep=0.5em,
      style=sameline,
      font=\normalfont,
      itemsep=3pt,
      topsep=4pt
    ]
    \item[\textbf{10} \textit{(Perfection):}] Reserved strictly for flawless execution with zero artifacts, hallucinations, or deviations.
    \item[\textbf{8} \textit{(Excellent):}] High-quality professional work with only negligible flaws visible upon close inspection.
    \item[\textbf{6} \textit{(Passable):}] The standard for ``average'' or ``okay'' generations.
    \item[\textbf{$<$ 5} \textit{(Flawed):}] Use these scores freely for any noticeable errors, distortions, or missed instructions.
  \end{description}

  \item \textbf{Missing References}: If a specific reference category is provided as an empty list \texttt{[]}, \texttt{None}, or a placeholder, you must set the score for that category to \textbf{$-$1} and the reasoning to ``N/A''.
  \item \textbf{Independence}: Evaluate each dimension independently. For example, a perfect Subject match can exist even if the Style is completely different (if that was the intent).
  \item \textbf{Output Format}: You must output strictly valid JSON without Markdown formatting.
  \item \textbf{Score Range}: Each score must be a float between 0 and 10.
  \item \textbf{Reasoning}: Provide a brief reasoning for each score.
\end{itemize}
\end{tcolorbox}
\captionof{figure}{System prompt for MLLM-based evaluation. The MLLM is required to provide both assessment scores and concise rationales to enhance reliability and facilitate debugging. Notably, a score of $-1$ is assigned to a reference type as a placeholder if the reference image is absent; such scores are excluded from the final score aggregation.}
\label{fig:mllm system prompt}

\begin{tcolorbox}[
  enhanced,
  breakable,
  colback=white,
  colframe=black!60,
  boxrule=0.6pt,
  arc=1pt,
  outer arc=1pt,
  toptitle=3pt,
  bottomtitle=3pt,
  coltitle=white,
  colbacktitle=headerGray,
  fonttitle=\bfseries\sffamily,
  title={User Prompt: MLLM-based Evaluation},
  left=8pt, right=8pt, top=6pt, bottom=6pt,
  fontupper=\small
]
\newpage
{\large\textbf{[User Prompt]}}
\medskip
\noindent\textbf{Evaluation Task}

\smallskip
\noindent Analyze the Target Generated Image against the inputs above. Follow these specific rules for each dimension:

\medskip
\noindent\textbf{1.\var{prompt\_following\_rubrics}}

\medskip
\noindent\textbf{2.\var{subject\_consistency\_rubrics}}

\medskip
\noindent\textbf{3.\var{style\_consistency\_rubrics}}

\medskip
\noindent\textbf{4.\var{lighting\_consistency\_rubrics}}

\medskip
\noindent\textbf{5.\var{background\_consistency\_rubrics}}

\medskip
\noindent\textbf{6.\var{aesthetic\_quality\_rubrics}}

\medskip
\noindent\textbf{Input Data}

\smallskip
\noindent Text Prompt: \var{text\_prompt}

\smallskip
\noindent Reference Images:
\begin{itemize}[leftmargin=1.5em, itemsep=1pt, topsep=2pt]
  \item Subject References: \var{subject\_ref\_label}
  \item Style Reference: \var{style\_ref\_label}
  \item Lighting Reference: \var{lighting\_ref\_label}
  \item Background Reference: \var{background\_ref\_label}
\end{itemize}

\noindent Target Generated Image: \var{generated\_img\_label}

\medskip
\noindent\textbf{Output Format}

\smallskip
\noindent Return a single JSON object. Do NOT use Markdown code blocks.

\smallskip
\noindent JSON Structure:

\begin{lstlisting}
{
  "text_adherence": {
    "score": <float>,
    "reasoning": "<string>"
  },
  "subject_consistency": {
    "score": <float>,
    "reasoning": "<string>"
  },
  ...
  "overall_average_score": <float>
}
\end{lstlisting}

\end{tcolorbox}
\captionof{figure}{User prompt for MLLM-based evaluation. These prompts incorporate granular scoring rubrics for each dimension, which appear as placeholders in this prompt, and are elaborated in \cref{fig:prompt following,fig:subject consistency,fig:style consistency,fig:lighting consistency,fig:background consistency,fig:aesthetic quality}.}
\label{fig:mllm user prompt}

\begin{tcolorbox}[
  enhanced,
  breakable,
  colback=white,
  colframe=black!60,
  boxrule=0.6pt,
  arc=1pt,
  outer arc=1pt,
  toptitle=3pt,
  bottomtitle=3pt,
  coltitle=white,
  colbacktitle=headerGray,
  fonttitle=\bfseries\sffamily,
  title={Evaluation Rubrics: Prompt Following},
  left=8pt,
  right=8pt,
  top=6pt,
  bottom=6pt,
  fontupper=\small
]

\newpage
{\large\textbf{[Prompt Following]}}
\medskip

\noindent\textit{Goal:} Verify if the generated image strictly follows the instructions in the Text Prompt, including the implicit or explicit instruction to utilize the provided Reference Images.

\smallskip
\noindent\textit{Focus:}
\begin{itemize}[leftmargin=1.5em, itemsep=1pt, topsep=2pt]
  \item \textbf{Entities \& Attributes:} Are the mentioned objects present with correct colors/shapes?
  \item \textbf{Actions \& Relations:} Are interactions and spatial positions correct?
  \item \textbf{Reference Integration (Crucial):} If the prompt requires using a reference (e.g., subject, style, background, or lighting), does the image respect those constraints?
\end{itemize}

\noindent\textbf{Critical Rules:}
\begin{enumerate}[leftmargin=1.5em, itemsep=1pt, topsep=2pt]
  \item A failure to match a Reference Image (Style, Subject, Background, Lighting) should result in a score lower than 6.
\end{enumerate}

\smallskip
\noindent\textit{Scoring} (0--10, \textbf{Conservative Scoring}):
\begin{itemize}[leftmargin=1.5em, itemsep=1pt, topsep=2pt]
  \item \textbf{0}: Completely irrelevant to the text prompt, or completely ignores the core reference requirement.
  \item \textbf{2}: The main subject is present, but the content or reference requirement is seriously violated.
  \item \textbf{4}:  The general content is partially captured, but important attributes, actions, or reference constraints are not properly followed.
  \item \textbf{6}:  The overall gist is correct. The subject, action, and setting generally match the prompt, but fidelity to the
requested references is weak or inconsistent.
  \item \textbf{8}: All major text instructions are satisfied, and the reference images are clearly utilized and recognizable. Only
minor deviations remain.
  \item \textbf{10}: Perfect alignment with both text and references. All required elements and constraints are accurately
integrated. A score of 10 should not be assigned if essential subject or style constraints are not preserved.\\
\end{itemize}

\end{tcolorbox}
\captionof{figure}{Evaluation rubrics for prompt following.}
\label{fig:prompt following}

\newpage
\begin{tcolorbox}[
  enhanced,
  breakable,
  colback=white,
  colframe=black!60,
  boxrule=0.6pt,
  arc=1pt,
  outer arc=1pt,
  toptitle=3pt,
  bottomtitle=3pt,
  coltitle=white,
  colbacktitle=headerGray,
  fonttitle=\bfseries\sffamily,
  title={Evaluation Rubrics: Subject Consistency},
  left=8pt,
  right=8pt,
  top=6pt,
  bottom=6pt,
  fontupper=\small
]

{\large\textbf{[Subject Consistency]}}
\medskip

\noindent\textit{Focus:}
Facial features, body or shape structure, distinctive markings, product geometry, logos, and characteristic design
details should be considered. Artistic style, rendering medium, lighting, and pose should be ignored.

\smallskip
\noindent\textit{Scoring} (0--10, \textbf{Conservative Scoring}):
\begin{itemize}[leftmargin=1.5em, itemsep=1pt, topsep=2pt]
  \item \textbf{0}: Wrong subject or complete absence. The target is a different person, animal, object, or product, or the intended subject is missing.

  \item \textbf{2}: Only a broad category match. The result belongs to the same general class, but the specific identity is not preserved, appearing generic rather than corresponding to the reference.
  \item \textbf{4}: The image attempts to preserve the subject, but important structural details are noticeably distorted or
inaccurate.
  \item ...
  \item \textbf{8}: Strong structural fidelity. Distinctive features are preserved with high accuracy, with only minor discrepancies under close inspection.
  \item \textbf{10}: Nearly perfect identity preservation, as if the same subject or object were reused in a different setting or style.\\
\end{itemize}

\end{tcolorbox}
\captionof{figure}{Evaluation rubrics for subject consistency.}
\label{fig:subject consistency}

\begin{tcolorbox}[
  enhanced,
  breakable,
  colback=white,
  colframe=black!60,
  boxrule=0.6pt,
  arc=1pt,
  outer arc=1pt,
  toptitle=3pt,
  bottomtitle=3pt,
  coltitle=white,
  colbacktitle=headerGray,
  fonttitle=\bfseries\sffamily,
  title={Evaluation Rubrics: Style Consistency},
  left=8pt,
  right=8pt,
  top=6pt,
  bottom=4pt,
  fontupper=\small
]

{\large\textbf{[Style Consistency]}}
\medskip

\noindent\textit{Goal:} Determine if the Target adopts the visual style of the Style Reference.

\noindent\textit{Focus:} Art medium, brushwork, color palette, texture, and overall visual atmosphere should be
considered, while the subject content should be ignored.

\smallskip
\noindent\textit{Scoring} (0--10, \textbf{Conservative Scoring}):
\begin{itemize}[leftmargin=1.5em, itemsep=1pt, topsep=2pt]
  \item \textbf{$-$1}: Reference missing / Not Applicable.
  \item \textbf{0}: Complete style mismatch.
  \item \textbf{2}: Only superficial resemblance in mood or color; the core medium or stylistic characteristics are incorrect.
  \item \textbf{4}: The broad style category is correct, but the specific stylistic signatures of the reference are not well captured.
  \item ...
  \item \textbf{8}: Strong stylistic match, including important texture, palette, and rendering characteristics.
  \item \textbf{10}: The target looks as if it were created by the same artist or medium.\\
\end{itemize}

\end{tcolorbox}
\captionof{figure}{Evaluation rubrics for style consistency.}
\label{fig:style consistency}

\newpage
\begin{tcolorbox}[
  enhanced,
  breakable,
  colback=white,
  colframe=black!60,
  boxrule=0.6pt,
  arc=1pt,
  outer arc=1pt,
  toptitle=3pt,
  bottomtitle=3pt,
  coltitle=white,
  colbacktitle=headerGray,
  fonttitle=\bfseries\sffamily,
  title={Evaluation Rubrics: Lighting Consistency},
  left=8pt,
  right=8pt,
  top=6pt,
  bottom=2pt,
  fontupper=\small
]

{\large\textbf{[Lighting Consistency]}}
\medskip

\noindent\textit{Goal:} Determine whether the target image transfers the lighting scheme of the Lighting Reference to the new scene defined by the prompt.

\noindent\textit{Focus:} Light direction, contrast, shadow quality, and color temperature should be considered. The lighting should be transferred to the new content rather than simply copying the original background.

\smallskip
\noindent\textit{Scoring} (0--10 Scale, \textbf{Conservative Scoring}):
\begin{itemize}[leftmargin=1.5em, itemsep=1pt, topsep=2pt]
  \item \textbf{$-$1}: Reference missing / Not Applicable.
  \item \textbf{0}: Lighting is completely inconsistent with the reference.
  \item \textbf{2}: Major failure, such as background leakage or a flat filter-like overlay, instead of true lighting transfer.
  \item \textbf{4}: The general atmosphere is vaguely similar, but the light direction, shadow behavior, or contrast is clearly incorrect.
  \item ...
  \item \textbf{8}: High-fidelity transfer of the lighting setup to the new scene.
  \item \textbf{10}: The lighting is integrated almost perfectly and behaves consistently with the new geometry.\\
\end{itemize}

\end{tcolorbox}
\captionof{figure}{Evaluation rubrics for lighting consistency.}
\label{fig:lighting consistency}

\begin{tcolorbox}[
  enhanced,
  breakable,
  colback=white,
  colframe=black!60,
  boxrule=0.6pt,
  arc=1pt,
  outer arc=1pt,
  toptitle=3pt,
  bottomtitle=3pt,
  coltitle=white,
  colbacktitle=headerGray,
  fonttitle=\bfseries\sffamily,
  title={Evaluation Rubrics: Background Consistency},
  left=8pt,
  right=8pt,
  top=6pt,
  bottom=4pt,
  fontupper=\small
]

{\large\textbf{[Background Consistency]}}
\medskip

\noindent\textit{Goal:} Determine whether the target image preserves the structural integrity and layout of the Background
Reference.

\noindent\textit{Focus:} Scene layout, object placement, perspective, and depth should be considered. If a style reference is also provided, texture or color changes are allowed, but the underlying scene structure should remain consistent.

\smallskip
\noindent\textit{Scoring} (0--10 Scale, \textbf{Conservative Scoring}):
\begin{itemize}[leftmargin=1.5em, itemsep=1pt, topsep=2pt]
  \item \textbf{$-$1}: Reference missing / Not Applicable.
  \item \textbf{0}: Completely unrelated background.
  \item \textbf{2}: Severe distortion, structural collapse, or an obvious pasted-together appearance.
  \item \textbf{4}: The background only matches the general theme or category, while the specific layout is not preserved.
  \item ...
  \item \textbf{8}: Major scene elements and layout are preserved with high fidelity.
  \item \textbf{10}: Near-perfect structural consistency, with no meaningful layout drift.
\end{itemize}

\end{tcolorbox}
\captionof{figure}{Evaluation rubrics for background consistency.}
\label{fig:background consistency}

\newpage
\begin{tcolorbox}[
  enhanced,
  breakable,
  colback=white,
  colframe=black!60,
  boxrule=0.6pt,
  arc=1pt,
  outer arc=1pt,
  toptitle=3pt,
  bottomtitle=3pt,
  coltitle=white,
  colbacktitle=headerGray,
  fonttitle=\bfseries\sffamily,
  title={Evaluation Rubrics: Aesthetic},
  left=8pt,
  right=8pt,
  top=6pt,
  bottom=6pt,
  fontupper=\small
]

{\large\textbf{[Aesthetic]}}
\medskip

\noindent\textit{Goal:} Evaluate the overall visual quality and appeal of the target image.

\noindent\textit{Focus:} Sharpness, composition, color harmony, realism or stylization quality, and the absence of visible artifacts should be considered. Evaluation should be made within the intended style.

\smallskip
\noindent\textit{Scoring} (0--10 Scale, \textbf{Conservative Scoring}):
\begin{itemize}[leftmargin=1.5em, itemsep=1pt, topsep=2pt]
  \item \textbf{0}: Unusable image with severe corruption.
  \item \textbf{2}: Major artifacts or broken anatomy.
  \item \textbf{4}: Technically usable, but visually unappealing or obviously low-quality.
  \item ...
  \item \textbf{8}: High-quality, polished image with strong visual appeal and no obvious artifacts.
  \item \textbf{10}: Exceptional visual impact, with outstanding technical and artistic quality.\\
\end{itemize}

\end{tcolorbox}
\captionof{figure}{Evaluation rubrics for aesthetic.}
\label{fig:aesthetic quality}